\documentclass[final]{siamltex}

\usepackage[]{fontenc}
\usepackage[latin1]{inputenc}
\usepackage{verbatim}
\usepackage{amsmath}
\usepackage{amssymb}
\usepackage{dsfont}
\usepackage{amsfonts}
\usepackage{mathrsfs}
\usepackage{graphicx}
\usepackage{color}
\usepackage{ulem}
\usepackage{xspace,bbm,epic,eepic}
\usepackage{pgfplots}
\usepackage{setspace}
\usepackage{xspace}

\usepackage{ stmaryrd }
\usepackage{graphicx}
\usepackage{lineno}
\usepackage{color}
\usepackage{url}
\usepackage{graphicx}
\usepackage{ stmaryrd }
\usepackage{algorithm}
\usepackage{algpseudocode}

\usepackage{amsmath,amssymb,color} % define this before the line numbering.
\newcommand{\s}{\mathbf{s}}

\newcommand{\Rr}{\mathds{R}}
\newcommand{\Cr}{\mathds{C}}

\newcommand{\thetaLearn}{\theta^{\text{learn}}}

\newcommand{\R}{P}

\newcommand{\YYY}{\mathbf{Y}} 
\newcommand{\AAA}{\boldsymbol{\Psi}_{\mathbf{X}}} 
\newcommand{\BBB}{\boldsymbol{\Psi}_{\mathbf{Y}}} 
\newcommand{\ZZZ}{\mathbf{Z}}

\newcommand{\TT}{T}
\newcommand{\TTPrim}{T_{\text{learn}}}

\newcommand{\NN}{N}
\newcommand{\NNPrim}{{N_{\text{learn}}}}
\newcommand{\eg}{\textit{e.g.}, }
\newcommand{\ie}{\textit{i.e.}, }

\DeclareMathOperator*{\argmax}{arg\,max}
\DeclareMathOperator*{\argmin}{arg\,min}

\newtheorem{constraint}{Constraint}

\begin{document}

\title{Non-linear Reduced Modeling of  Dynamical Systems using kernel methods and low-rank approximation }

\author{P. H\'EAS$^{\dagger,}$\thanks{INRIA, $^\dagger$IRMAR, $^\sharp$IRISA,  Univ. Rennes, 263 Av. G\'en\'eral Leclerc, Campus de Beaulieu, 35042 Rennes, France ({\tt patrick.heas@inria.fr,cedric.herzet@inria.fr,benoit.combes@inria.fr}) {\hspace{+4cm}This paper is an improved and extended version of the conference proceeding~\cite{HeasIcassp2020}.}}  \and C. HERZET$^{\dagger,*}$ \and   B. COMB\`ES$^{\sharp,*}$
   }

\date{}
\maketitle

\begin{abstract}
Reduced modeling of a computationally demanding dynamical system aims at approximating
its   trajectories, while optimizing the trade-off between accuracy and
computational complexity. In this work, we propose to   achieve such  an approximation by first embedding the trajectories in a reproducing kernel Hilbert
space (RKHS), which has interesting approximation and calculation capabilities, and then solving the associated reduced model problem. More specifically, we propose a new efficient algorithm for
data-driven reduced modeling of non-linear dynamics based on linear
approximations in a RKHS. This algorithm takes advantage of the closed-form
solution of a low-rank constraint optimization problem while exploiting
advantageously kernel-based computations. Reduced modeling with this algorithm
reveals a gain in approximation accuracy, as shown by numerical simulations,
and in  complexity with respect to existing approaches.
\end{abstract}
\begin{keywords}
 Reduced modeling, kernel methods,   low-rank approximations,  non-linear dynamics \vspace{-0.25cm}
\end{keywords}

\section{Introduction} \vspace{-0.cm}

Consider  a high-dimensional system  {of the form:} \vspace{-0.cm}
\begin{align}\label{eq:model_init} 
 \left\{\begin{aligned}
& x_{t}(\theta)= f_t(x_{t-1}(\theta)) , \quad t=2,\ldots,\TT,\\
&x_1(\theta)={\theta},
\end{aligned}\right. \vspace{-0.6cm}%\\
\end{align} 
\noindent
 {where}  the $x_t$'s and $\theta$ in  $\Rr^p$, constitute respectively a state trajectory and an initial condition, and where $f_t:\Rr^p \to \Rr^p$ is an arbitrary function whose direct evaluation is considered to be time consuming when $p$ is large. 
Reduced modeling of the dynamical systems aims to lighten the computation load for the evaluation of a trajectory. The idea common to all model reduction methods is the projection of the  high-dimensional system \eqref{eq:model_init} onto a  low-dimensional manifold, which enables a rapid and reliable approximation of a trajectory for any initial condition.
Projections onto well-chosen linear subspaces have been extensively studied in the reduced modeling literature, leading in particular to the well-established reduced basis methods~\cite{quarteroni2015reduced}.  Data-driven approaches have  emerged for the case where the function $f_t$ is unknown and we only have access to a set of representative solutions of \eqref{eq:model_init}. 
{\it Dynamic mode decomposition} (DMD)~\cite{Schmid10} is a popular   framework for this purpose.  
It consists in the data-driven linear approximation of the function $f_t$: a  matrix substituting $f_t$ is learned  from representative trajectories, so-called \textit{snapshots},  by minimizing the norm of the residual  of   linear approximations subject to  specific constraints \cite{Chen12,heas2017optimal,HeasHerzet17,Jovanovic12}. \medskip%The  eigen decomposition of the minimizer constitutes then  the core of the reduced model.
  
In recent decades, the community's interest has shifted to nonlinear reduced modeling, in particular, projections onto time-dependent subspaces~\cite{billaud2017dynamical,cagniart2019model,koch2007dynamical,nouy2010proper} or approaches based on nonlinear embeddings of the solution manifold in a space with better  approximation capabilities~\cite{iollo2014advection,taddei2020registration}.  Among the latter  approaches, the data-driven DMD  has been extended\footnote{We mention that there are other nonlinear extensions of DMD involving parametric regression which have been proposed for reduced modeling of dynamical systems that are polynomial~\cite{peherstorfer2016data} or in analytic form~\cite{benner2020operator}. Such approaches are interesting but beyond the scope of this article, which focuses on non-parametric approaches based on nonlinear embeddings.} to a method often referred to as {\it extended DMD} (EDMD) \cite{bouvrie2017kernel,giannakis2023learning,Lusch2018DeepLF,williams2015data,williams2014kernel,yeung2019learning}.  Basically,  DMD and EDMD  are identical,  except that EDMD first  immerses  the trajectory through a non-linear mapping $\Psi$ in a space exhibiting better low-rank approximation capabilities.   
More explicitly,
let $\Psi: \Rr^p \to \mathcal{H}$, where  $\mathcal{H}$  is a Hilbert space  endowed with the inner product  $\langle \cdot ,\cdot \rangle_{\mathcal{H}}$  and the induced norm $\| \cdot \|_\mathcal{H}$. 
EDMD approximates system \eqref{eq:model_init} through a system taking the form of \vspace{-0.1cm}
 \begin{align}\label{eq:model_koopman_approxObservable0} 
 \left\{\begin{aligned}
&\eta_t(\theta)= \hat A_k \eta_{t-1}(\theta), \quad t=2,\ldots,{\TT},\\
&\eta_1(\theta)=\Psi({\theta}),
\end{aligned}\right. \vspace{-0.5cm}%\\
\end{align} 
where $\hat A_k : \mathcal{H} \to \mathcal{H}$ is a  linear  operator   of rank at most  $k$ (satisfying some optimality criterion which will be specified later),  yielding an approximation of the state $x_{\TT}(\theta)$ by  an inverse mapping \vspace{-0.1cm}
\begin{align}\label{eq:model_koopman_approxObservable} 
\tilde x_{{\TT}}(\theta)=\Psi^{-1}(\eta_{\TT}(\theta)).\vspace{0.1cm}
\end{align}
A proper definition of the inverse map $\Psi^{-1}$  will be given in the next section.\medskip

This paper focuses on reduced models of the form  \eqref{eq:model_koopman_approxObservable0}-\eqref{eq:model_koopman_approxObservable},   where   $\dim(\mathcal{H})\gg p$ (including $\dim(\mathcal{H})=\infty$). Such  embeddings are appealing due to the ability of high-dimensional Hilbert spaces to linearize   differential equations~\cite{koopman1931hamiltonian,kowalski1991nonlinear,mezic2004comparison}. 
%  \textcolor{red}{[
% pour completer cet argument 
% \begin{itemize}
% \item parler brievement du probleme dans un formalisme proba que revient a  estimer spectre noyau markovien et que travaux recents (Klus2019,das2012, etc) montre que leurs approches d'estimation data-based  de l operateur de transition (Koopman)  converge qd m tend vers $\infty$
% \item Rajouter que la convergence de notre algo a priori a toutes les raisons de pouvoir etre etudie, mais est hors du scope du papier et on laisse comme question ouverte pour d'autres, car tres compliques!. En revanche empiriquement on prouve que bonne performance pour $m$ fini par rapport etat de l art
% \item dire que point faible c est que l'on ne sait pas a priori pour quelle classe de dynamique une transformation $\Psi$ de l espace d etat fera  qu'il existe un $A_k$  de bas rang qiu soit exact ou du moins une bonne approximation (plus que dans l espace original)...En revanche empiriquement on prouve que bonne performance + on est capable de calculer exactement l erreur d'apprentissage (section 3.3.2)
% \end{itemize}
% ]}
  Nevertheless, computing a reduced model using these high-dimensional embeddings is  difficult  since neither the inference of operator  $\hat A_k$ nor the recursion  \eqref{eq:model_koopman_approxObservable0} or the inversion \eqref{eq:model_koopman_approxObservable} is   tractable in a general setting. 
% Computing a reduced model using these high-dimensional embeddings is obviously difficult  since neither the inference of operator  $\hat A_k$ nor the recursion  \eqref{eq:model_koopman_approxObservable0} or the inversion \eqref{eq:model_koopman_approxObservable} is   tractable in a general setting. 
 To obtain a ``good'' trade-off between accuracy and complexity of the reduced model,  one needs to accomplish two challenging  tasks: {\it i)} learn a tractable representation of a low-rank operator $\hat A_k$ yielding an accurate approximation of the form \eqref{eq:model_koopman_approxObservable0}-\eqref{eq:model_koopman_approxObservable}, {\it ii)}   build a  low-complexity algorithm able to compute  $\tilde x_{\TT}(\theta)$ satisfying  \eqref{eq:model_koopman_approxObservable0}-\eqref{eq:model_koopman_approxObservable} for a given $\theta$ and $T$.\medskip

State-of-the-art methods such as~\cite{Chen12,HeasHerzet17,hemati2017biasing,Jovanovic12,Tu2014391} involve a complexity in $\dim(\mathcal{H})$, thus are  non-efficient in  high-dimensional settings. Alternatively, authors in \cite{williams2014kernel} have introduced an efficient  algorithm to compute \eqref{eq:model_koopman_approxObservable0} - \eqref{eq:model_koopman_approxObservable}  for any map $\Psi$ related to a reproducing kernel Hilbert space (RKHS) \cite{steinwart2006explicit}.
% The idea is to rely  on the {\it kernel trick}~\cite{bishop2006pattern} to compute recursion   \eqref{eq:model_koopman_approxObservable0} and build  approximation   \eqref{eq:model_koopman_approxObservable}. 
  This algorithm known as {\it kernel-based DMD} (K-DMD) enjoys  an advantageous complexity  linear in $p$ and independent of $\dim(\mathcal{H})$ and of the trajectory length $T$. Unfortunately, it relies on a set of restrictive assumptions.   The regularized kernel regression technique~\cite{scholkopf1998nonlinear} known as {\it kernel analog forecasting}~\cite{alexander2020operator,giannakis2023learning} can be seen as the K-DMD algorithm, in the particular case where the reduced model is trained to compute only \eqref{eq:model_koopman_approxObservable} for some specific and predefined path length $T$.   In the article~\cite{bouvrie2017kernel}, the authors propose a kernel method similar to K-DMD\footnote{It should be noted that the method~\cite{bouvrie2017kernel}  applies to a more general class of models, namely non-linear control systems, which can be identified with system~\eqref{eq:model_init} in the particular case where the observation operator is the identity}, but alleviating some of its restrictive assumptions. However, it should be noted that the method involves additional approximation steps using regularized kernel regression techniques. In addition, the method produces a reduced model whose complexity is no longer independent of $T$.    \medskip
  
The contribution of this work is to propose a new algorithm dubbed ``generalized kernel-based DMD (GK-DMD)'' that generalizes   K-DMD to less restrictive assumptions, while being characterized by a gain in computational complexity and approximation accuracy, as evidenced by our numerical simulations. To put this work into context, it is important to make the following two points: {\it i)} although DMD-type approximations in RKHS have recently attracted the interest of several research groups~\cite{das2021reproducing,kawahara2016dynamic,klus2020eigendecompositions,korda2018convergence}, a theoretical characterization of the convergence of the proposed approximation is out of the scope of our paper; this paper is limited to providing a methodology and empirical evidence of its  performances   through relevant numerical simulations;
{\it ii)} in contrast to most of the recent studies concerned with the Koopman operator approximation, the paper focuses instead on the approximation the trajectories of $\eqref{eq:model_init}$ in the ambient space, by means of learning  a low-rank operator  in a RKHS, in the line of the kernel method proposed in~\cite{bouvrie2017kernel} and of  previous results obtained in \cite{HeasHerzet18Maps,HeasHerzet17}.\medskip % this involves computing an approximation  subject to a low-rank constraint of the dynamics mapped  in the RKHS  $\mathcal{H}$, the  low-rank approximation in $\mathcal{H}$ is then used to construct a non-linear reduced model of the trajectory mapped back in the ambient space  $\Rr^p$. 

   %In particular, we show how to use the optimal  (and possibly infinite-dimensional) map $A_k^\star$ given in~\eqref{eq:prob1} in order  to compute exactly the approximation~\eqref{eq:model_koopman_approxObservable} with a complexity linear in $p$ and $k$ and independent of $\dim(\mathcal{H})$ and ${\TT}$. 
 
 %This independence in both, the Hilbert space dimension and the trajectory length, relies on the {kernel trick} and on the diagonalisation of recursion~\eqref{eq:model_koopman_approxObservable0}.   In a nutshell, the kernel trick is used to first recast  recursion~\eqref{eq:model_koopman_approxObservable0} as a linear combination of inner products with  eigen-vectors of $A_k^\star$. It is then used once more for the computation of the inverse mapping from $\mathcal{H}$ to $\Rr^p$ and the obtention of the solution~\eqref{eq:model_koopman_approxObservable}. Numerical simulations demonstrate that the  solution computed by GK-DMD lowers significantly the approximation error in comparison to state-of-the-art techniques, especially in the case where $k <N(T-1)$, which is of interest for reduced modeling. Moreover, the numerical evaluation reveals that the proposed method is robust to noise and   overfitting.

The paper is organized as follows. Section~\ref{sec:StateOfTheArt}  specifies the reduced modeling problem of interest and reviews existing solutions. Our generalized kernel-based algorithm is  presented  in  Section~\ref{sec:okDMD}. The proposed algorithm is  evaluated numerically in Section~\ref{sec:numEval}. Proofs of theoretical results and technical details are provided in the appendices.

\section{Problem and Existing Solutions}\label{sec:StateOfTheArt}\vspace{-0.cm}

\subsection{The reduced modeling problem}\label{sec:ReducedModelingProb}
In this work,   we  are interested in the design of an  algorithm evaluating for any $\theta \in \Rr^p$ the approximation $\tilde x_{\TT}(\theta)$  given by   \eqref{eq:model_koopman_approxObservable0}-\eqref{eq:model_koopman_approxObservable}, with a complexity  constant with respect to the  dimension $\dim(\mathcal{H})$ and the trajectory length $\TT$. Moreover, we are interested in a data-driven approach: the algorithm should rely on a  reduced model  learned from a set of representative trajectories $\{x_t(\thetaLearn_i)\}_{t=1,i=1}^{\TTPrim,\NNPrim}$ of the high-dimensional system \eqref{eq:model_init}  corresponding to  $\NNPrim$ initial conditions $\{ \thetaLearn_i\}_{i=1}^\NNPrim$ and where $\TTPrim$ is possibly different from $\TT$.  We  thus consider a particular instance of the family of data-driven reduced model given by   \eqref{eq:model_koopman_approxObservable0}-\eqref{eq:model_koopman_approxObservable}, where the low-rank operator $\hat A_k$ and  the inverse map $\Psi^{-1}$ have  specific definitions.\medskip

In order to introduce these definitions,  we  need   some notations. The function space $\mathcal{B}(\mathcal{V},\mathcal{U})$ will be the set of linear bounded operators from $\mathcal{V}$ to  $\mathcal{U}$. The subset of low-rank operators will be denoted 
$$\mathcal{B}_k(\mathcal{V},\mathcal{U}) = \{  M \in \mathcal{B}(\mathcal{V},\mathcal{U}): \textrm{rank}(M)\le k\}.$$
  The adjoint of an operator  $M \in \mathcal{B}(\mathcal{V},\mathcal{U})$ will be denoted $M^* \in  \mathcal{B}(\mathcal{U},\mathcal{V})$. The Hilbert-Schmidt norm  $\|\cdot\|_{\mathcal{HS}}$  of  $M \in \mathcal{B}(\mathcal{V},\mathcal{U}) $ is  defined as
$\|M \|^2_{\mathcal{HS}}=\sum_{i=1}^{\dim(\mathcal{V})} \| M e_i\|^2_{\mathcal{U}},$ where $\{e_i\}_{i=1}^{\dim(\mathcal{V})}$ is an orthonormal basis of $\mathcal{V}$. The $\ell_2$-norm of a vector  $v\in \mathcal{V}$, with $\dim(\mathcal{V})<\infty$, will be denoted  $\|v \|_2$.
For a  vector $v\in \mathcal{V}$, the subscript notation $v_i$ will denote its  $i$th component.    The distinction from the subscript notation $\thetaLearn_i$  denoting the $i$th element of a set $\{\thetaLearn_i\}_{i>0}$ will be  clear from the context.\medskip

%
%
%Using the training data, inspired by the low-rank approximations  introduced in~\cite{Chen12,Jovanovic12}, we consider a generalization of the minimization problem in \cite{williams2015data,williams2014kernel}.   The low-rank operator $\hat A_k$ is set as the solution of the constrained optimization problem
%

{\bf Low-rank operator}. The first definition concerns the low-rank operator  learned from the training data set $\{x_{t}(\thetaLearn_i)\}_{t=1,i=1}^{\TTPrim,\NNPrim}$.
%Assume that for learning the reduced model,  we have at our disposal   $i=1,\ldots,N$ representative initial conditions $\thetaLearn_i$ and  the related  trajectories  ${x_1(\thetaLearn_i),\ldots,x_{\TTPrim}(\thetaLearn_i)}$   computed using system \eqref{eq:model_init} usually called snapshots, which are   
%of length $\TTPrim$, with $\TTPrim$ possibly different from $T$.  
Let us define   operators $\AAA,\, \BBB \in \mathcal{B}(\Rr^m,\mathcal{H}) $, with $m=\NNPrim(\TTPrim-1),$  for any $w \in  \Rr^m$ as the linear combinations \vspace{-0.1cm}
$$
 \AAA w = \sum_{i,t=1}^{\NNPrim,\TTPrim-1} \Psi(x_{t}(\thetaLearn_i)) w_{\ell(i,t)}\quad  \textrm{and}\quad  
 \BBB w=  \sum_{i,t=1}^{\NNPrim,\TTPrim-1} \Psi(x_{t+1}(\thetaLearn_i)) w_{\ell(i,t)},\vspace{-0.1cm}$$
 where $\ell(i,t)={(\TTPrim-1)(i-1)+t}.$
 Using these operators, we consider  the constrained optimization problem\vspace{-0.1cm}
\begin{align}\label{eq:prob1} 
		A_k^\star \in &\argmin_{A\in \mathcal{B}_k(\mathcal{H},\mathcal{H})} \|\BBB -A  \AAA \|_{\mathcal{HS}}.
		\end{align} 
We remark that the square of the Hilbert-Schmidt norm  in \eqref{eq:prob1} can be rewritten as a sum  of square norms of the form  
$\|\Psi(x_{t+1}(\thetaLearn_i))-  A\Psi(x_{t}(\thetaLearn_i))\|^2_{\mathcal{H}}$ over $t=1, \ldots, \TTPrim-1$ and $i=1,\ldots,\NNPrim$. Hence, the objective function in \eqref{eq:prob1}  measures the  overall discrepancy between the true states and their prediction by operator $A$ from the previous states, mapped into the Hilbert space. \medskip

 \begin{definition}\label{property:1}
     The low-rank operator $\hat A_k$ in \eqref{eq:model_koopman_approxObservable0} is defined as the solution of~\eqref{eq:prob1}.\medskip
 \end{definition}

 This first definition stipulates that the learned low-rank linear operator for reduced modeling in $\mathcal{H}$ is optimal in the Hilbert-Schmidt norm. Note that  \eqref{eq:prob1} is a generalization of the solution of the minimization problem in \cite{williams2015data,williams2014kernel}, subject to a low-rank constraint as in~\cite{Chen12,Jovanovic12}.  Indeed, let $ \hat A^{\ell s}_k$ be defined as a $k$-term truncation of the eigendecomposition of the solution of the unconstrained problem  $\BBB  \AAA^\dagger=\argmin_{A\in \mathcal{B}(\mathcal{H},\mathcal{H})} \|\BBB -A  \AAA  \|_{\mathcal{HS}}$~\cite{Tu2014391}.
Then $ \hat A^{\ell s}_k$ differs from the optimal solution $A_k^\star$ for $k<m$, while for $k\ge m$, both operators are equivalent.\medskip 

%A closed-form solution  $A_k^\star$ of  problem \eqref{eq:prob1} is available for the case where $\dim(\mathcal{H}) < \infty$  \cite{HeasHerzet17}  and for the general case of infinite-dimensional Hilbert spaces~\cite{HeasHerzet18Maps}. 

{\bf Minimum distance estimation}. The second definition concerns the inverse map. \medskip \vspace{-0.cm}%\footnote{The definition and the existence of the inverse \eqref{eq:defInverse2} is discussed in Appendix~\ref{app:1}.}   

 \begin{definition}\label{property:2}
    The inverse map $\Psi^{-1}$ from $\mathcal{H}$ to $\Rr^p$ in \eqref{eq:model_koopman_approxObservable} is defined as\begin{align}\label{eq:defInverse2}
 \Psi^{-1}(\eta) &\in
\arg\min_{z \in  \Rr^p}\|{ \eta}-{\Psi(z)}\|_{\mathcal{H}}.\vspace{-0.cm}
\end{align}\medskip
 \end{definition}
 
 This second definition stipulates that the inverse mapping  is optimal in a minimum distance sense. This definition is further discussed in Appendix~\ref{app:1.1}.\medskip

{\bf Low complexity}.\,\,In addition, we require the algorithm evaluating the approximation  $\tilde x_{\TT}(\theta)$  to be a low-complexity one in the following sense.\medskip
  
   \begin{constraint}\label{property:3}
     The complexity for evaluating $\tilde x_{\TT}(\theta)$ given by   \eqref{eq:model_koopman_approxObservable0}-\eqref{eq:model_koopman_approxObservable}  is  constant with respect to  $\dim(\mathcal{H})$ and~$\TT$.\\
 \end{constraint}

In order to enable a complexity constant  in ${\TT}$, we will assume all along this work  that $\mathcal{H}$ is separable and that $A_k^\star $ is  diagonalizable\footnote{ This assumption holds in particular if the linear bounded  operator $A_k^\star $ is  compact  self-adjoint or normal \cite{zhuoperator}, or in the case where $\dim(\mathcal{H}) <\infty$  and all non-zero eigen-values are distinct~\cite{Horn12}.}.  These  assumptions enable  to evaluate  recursion~\eqref{eq:model_koopman_approxObservable0}  independently of the trajectory length ${\TT}$. 
Explicitly, let  $\{\xi_i\}_{i\in \mathbb{N}}$ and   $\{\zeta_i\}_{i\in \mathbb{N}}$ be   bases of $\mathcal{H}$ associated to the left and right eigen-vectors of  $A_k^\star $, \ie $A_k^\star  \zeta_i = \lambda_i \zeta_i$ and  $(A_k^\star)^* \xi_i   = \lambda_i \xi_i$  for $i=1,\ldots, \dim(\mathcal{H})$, where $\{\lambda_i\}_{i}$ is the related sequence of eigen-values sorted by decreasing magnitude. The finite rank of operator $A_k^\star $ (which has at most $\textrm{rank}(A_k^\star )\le k$ non-zero eigen-values) and the bi-orthogonality of the left and right eigen-vectors yield  
%\begin{align}\label{eq:eigenAk
$A_k^\star  \Psi = \sum_{i=1}^{k} \lambda_i\langle\xi_i, \Psi \rangle_{\mathcal{H}}   \zeta_i.$
%\end{align}%The obtention of this alternative formulation is also detailed in the  appendix. 
%The columns of matrices
%  $(\Xi^{-1})^*=(\xi_1\cdots \xi_n)\in \Cr^{n \times n} $ and  $\Xi=(\zeta_1\cdots \zeta_n)\in \Cr^{n \times n} $   are left and right eigen-vectors  associated to  the Jordan matrix  $\Lambda \in \Cr^{n \times n}$ of rank at most $k$~\cite{golub2013matrix}.
%Assuming that $\hat {A}_k$ is diagonalisable, then  $\Lambda =\textrm{diag}(\lambda_1,\ldots, \lambda_n)$, and 
Defining   $  \varphi_i(\theta) = \langle \xi_i, \Psi(\theta) \rangle_{\mathcal{H}}$, \eqref{eq:model_koopman_approxObservable} then  becomes \vspace{-0.1cm}
\begin{align}\label{eq:koopman1}
 \tilde x_{\TT}(\theta)&=\Psi^{-1} ( \sum_{i=1}^{k}\nu_{i,{\TT}}\zeta_i ),\quad 
\nu_{i,{\TT}} =  \lambda_i^{{\TT}-1}  \varphi_i(\theta).\vspace{-0.4cm}
\end{align}  % and where the $ \xi_i$'s and the $\zeta_i$'s are associated to the  non-zero eigen-values $\lambda_i \in \Cr$  of $\hat {A}_k$. 
% Note that if  $\Psi^{-1}$ is linear, then \eqref{eq:koopman1} simplifies to \vspace{-0.cm}
%\begin{align}\label{eq:eigenmoderepre}
% \tilde x_{{\TT}}(\theta)= \sum_{i=1}^{k} \nu_{i,{\TT}} \mu_i, \quad\textrm{with}\quad  \mu_i=  \Psi^{-1} \zeta_i \in \Cr^p.\vspace{-0.15cm}
% \end{align}
%From \eqref{eq:koopman1} or \eqref{eq:eigenmoderepre}, it is now clear that an eigen diagonalization of $A_k^\star $ enables  the computation of approximation  $\tilde x_{\TT}(\theta)$  with a complexity independent of ${\TT}$.
%We mention that  $ \varphi_i$'s,  $\mu_i$'s  and  $\lambda_i$'s are known in the literature  respectively as approximations of  the $i$th Koopman's {\it  eigen-function}, {\it  eigen-mode} and {\it eigen-value} \cite{williams2015data}. 
Note that in the case where  $\Psi^{-1}$ is linear, then \eqref{eq:koopman1} simplifies to \vspace{-0.2cm}
\begin{align}\label{eq:eigenmoderepre}
 \tilde x_{{\TT}}(\theta)= \sum_{i=1}^{k} \nu_{i,{\TT}} \mu_i, \quad\textrm{with}\quad  \mu_i=  \Psi^{-1} \zeta_i \in \Cr^p.\vspace{-0.15cm}
 \end{align}
We mention that,   under specific assumptions, $ \varphi_i$'s,  $\mu_i$'s  and  $\lambda_i$'s  may be convergent approximations of the $i$th Koopman's {\it  eigen-function}, {\it  eigen-mode} and {\it eigen-value} \cite{das2021reproducing,korda2018convergence}. However, this issue is beyond the scope of this paper.\medskip

In summary, the problem is to
design an algorithm computing for any  $\theta\in\Rr^p$ the approximation  $\tilde x_{\TT}(\theta)$  given by   \eqref{eq:model_koopman_approxObservable0}-\eqref{eq:model_koopman_approxObservable} using Definitions \ref{property:1} and \ref{property:2}, with a complexity satisfying Constraint \ref{property:3}.

\subsection{Existing solutions}
In this section, we discuss  existing methods that can be used to partially solve the problem  described in Section~\ref{sec:ReducedModelingProb}. They  will serve as   ingredients for our GK-DMD algorithm.
To introduce these state-of-the-art methods, we  need to introduce the notations used for SVDs, pseudo-inverses and orthogonal projectors.\medskip

First, the SVD  of $M \in \mathcal{B}_k(\mathcal{V},\mathcal{U})$  will be denoted
$M  =\sum_{i=1}^{k} \sigma_i^M  u_i^M \langle v_i^M, \cdot \rangle_{\mathcal{V}} ,$
where $\{v^M_i\}_{i=1}^{k}$  are respectively the left and right singular vectors associated to the sequence of decreasing  singular values  $\{\sigma^M_i\}_{i=1}^{k}$  of $M$.  We will use  the short-hand SVD notations: $$M  =  U_M  \Sigma_M  V_M^*,$$ where $U_M\in \mathcal{B}(\Cr^k,\mathcal{U})$,  $\Sigma_M\in \mathcal{B}(\Cr^k,\Cr^k)$ and $ V_M^*\in \mathcal{B}(\mathcal{V},\Cr^k)$ are defined 
for any vector $w\in  \mathcal{V},\, s \in \Cr^k$ as
$
 U_M s = \sum_{j=1}^{k} u_j^M s_{j}, \quad  (V_M w)_i = \langle v_i^M, w \rangle_{\mathcal{V}}\quad \textrm{and} \quad
( \Sigma_M s)_i=  \sigma_i^M s_i.
 $
Then, the pseudo-inverse of $M$ denoted $ M^\dagger \in \mathcal{B}(\mathcal{U},\mathcal{V})$  is  defined as
$$M^\dagger  =\sum_{i=1}^{m} (\sigma_i^M)^{\dagger}  v_i^M \langle u_i^M, \cdot \rangle_{\mathcal{U}} ,\quad \textrm{where}\quad 
(\sigma_i^M)^{\dagger}=  \left\{\begin{aligned}
&(\sigma_i^M)^{-1}\quad \textrm{if}\quad  \sigma_i^M > 0\\%\mbox{\quad$\forall\, s\in\Sc,t\in\Tc$},
&0\quad\quad\quad \textrm{else}
\end{aligned}\right. \vspace{-0.cm}%\\
.$$  Its SVD will be written in short-hand notations as   $M^\dagger  =  V_M  \Sigma^\dagger_M  U_M^*.$
Finally, the orthogonal projector onto the image of $M$ (resp. of $M^*$) will be denoted by\cite{golub2013matrix} $$\mathbb{P}_{M}=M   M^\dagger \quad \textrm{(resp. } \mathbb{P}_{M^*}=M^\dagger  M\textrm{)}.$$
We are now in a position to present the two main classes of approaches that exist.
\vspace{-0.1cm}\\
%The first ingredient is a method computing the desired reduced model but which becomes intractable in the case where $\dim(\mathcal{H})$ is too large. The second ingredient is a tractable reduced model for any value of $\dim(\mathcal{H})$  which relies on restrictive assumptions. 

\textbf{Optimal but intractable}. Reduced model \eqref{eq:koopman1} where $A_k^\star $ is replaced by $ \hat A^{\ell s}_k$, \ie by a truncation to $k$ terms of the eigendecomposition of $A_m^\star=\BBB  \AAA^\dagger $, is called {EDMD}.   The reduced model \eqref{eq:koopman1} with $A_k^\star $ given by \eqref{eq:prob1} is called \textit{low-rank} EDMD. As in general for $k<m$, $A_k^\star \neq \hat A^{\ell s}_k$, the two reduced models differ, while for $k\ge m$ $ A^\star_k=\hat A^{\ell s}_k=A_m^\star.$  As we wish to be in agreement with  Definition~\ref{property:1} in any setting, and typically for $k<m$, we focus on  {low-rank EDMD}  and the characterization of  operator $A_k^\star$.  \medskip% In the case where  $\dim(\mathcal{H}) < \infty$,  \cite[Theorem 4.1]{HeasHerzet17}  provides the  solution $A_k^\star$ of  problem \eqref{eq:prob1} for an arbitrary value of $k$.  

A closed-form expression for operator $A_k^\star$ is available using the generalization of our result in \cite[Theorem 4.1]{HeasHerzet17}   to  separable infinite-dimensional Hilbert spaces~\cite[Theorem 2.1]{HeasHerzet18Maps}.
 The latter theorem yields a solution\footnote{Note that the solution  $A_k^\star$ is not  unique when $\boldsymbol{\Psi}_\mathbf{X}$ is not full rank. Indeed, there exist an infinity of solutions  of \eqref{eq:prob1}  of the form $A_k^\star+M$, where the columns  of $V_M$ are orthogonal to those  of $U_{\boldsymbol{\Psi}_\mathbf{X}}$.} of  problem \eqref{eq:prob1} for arbitrary value of $k$ \vspace{-0.15cm}
\begin{align}\label{eq:solAkOpt}
 A^\star_k=\mathbb{P}_{\ZZZ^k}\BBB \AAA^\dagger,\vspace{-0.15cm}
\end{align}
with the orthogonal projector $\mathbb{P}_{\ZZZ^k}=\hat \R_k \hat \R_k^*$ and \vspace{-0.15cm}
\begin{align}\label{eq:Pk}
\hat \R_k \in  \mathcal{B}(\Rr^k, \mathcal{H}):  w \to  \sum_{i=1}^k u^{\ZZZ}_i w_i,\vspace{-0.2cm}
\end{align}
 where singular vectors $u^{\ZZZ}_i$'s are issued from the SVD (see notations above) of   \vspace{-0.15cm}
  \begin{align}
  \label{eq:Z} 
  \ZZZ= \BBB \mathbb{P}_ {\AAA^*}\in \mathcal{B}(\Rr^m, \mathcal{H}).\vspace{-0.1cm}
  \end{align}
However, to meet Definition~\ref{property:2} and achieve Constraint~\ref{property:3}, it remains to propose a low-complexity algorithm able to build and evaluate reduced model  \eqref{eq:koopman1} from the closed-form,  but potentially infinite-dimensional,  solution $A_k^\star$.
Interestingly,  the projector $\mathbb{P}_{\ZZZ^k}$ in the closed-form solution  \eqref{eq:solAkOpt}   implies that $A_k^\star \eta$, for any $\eta \in \mathcal{H}$, belongs to the span of operator $\Psi_\YYY$. Therefore, an algorithm able to compute~\eqref{eq:koopman1} by evaluation of scalar products of the form $\langle A_k^\star \eta,\Psi(z) \rangle_{\mathcal{H}}$, where  $z \in \Rr^p$, will have a complexity linear in the cost of the evaluation of a scalar product of the form  $\langle \Psi(y_i), \Psi(z) \rangle_{\mathcal{H}}$, where the  elements in the set $\{x_{t+1}(\thetaLearn_i)\}_{t=1,i=1}^{\TTPrim-1,\NNPrim}$ define the columns $\{y_i\}_{i=1}^{m}$ of matrix $\YYY\in \Rr^{p \times m}$.  This idea constitutes the grounds of our algorithm presented in  Section~\ref{sec:algo}.\vspace{-0.1cm}\\

\textbf{Tractable but restrictive.}
%The remaining question is how  to build and evaluate reduced model of the form of \eqref{eq:koopman1}  for any $\theta$ in some range of interest  with a low complexity independent of $\dim(\mathcal{H})$?   
To tackle the high-dimensional setting  $\dim(\mathcal{H})~\gg~p$ related to  an EDMD reduced model,  authors of  the seminal work~\cite{williams2014kernel}   consider     a specific class of  mapping suited for the case where $\dim(\mathcal{H})~\gg~m$:
%\textcolor{red}{[\begin{itemize}
%\item Ajouter approches identiques de Kawahara NIPS2016  a une projection dans un autre sous-espace pres   ou de  Schwantes2021 qui introduit une regularisation, tout ca pour calculer $\Psi_\XXX^{-1}$ mal defini, sans pseudio inverse $\Psi_\XXX^\dagger$, 
%\item parler des approches identique de Klus2019 (qui regularise aussi) ou similaire de Das2021 (qui utilise un "generateur regularise" ...  chaud a comprendre !!), qui prouve  la convergence vers le spectre d un noyau markovien continue (operateur koopman) quand probleme d approximation formule en proba   lorsque  m tend vers $\infty$ 
%\item Insister sur l approche de modele reduit a notre conaissance jamaios abouti dans etat l art, on calcul la prediction dans le RKHS (les observables), on s interesse pas a la prediction dans l espace originel.
%\end{itemize}]}.
 the mapping $\Psi$ is such that a scalar product in ${\mathcal{H}}$ of the form $\langle \Psi(z), \Psi(y) \rangle_{\mathcal{H}}$ with $z, y \in \Rr^p$ is given by the evaluation of 
a symmetric positive definite kernel   \vspace{-0.1cm}
$$
h\,:\,\Rr^p \times \Rr^p \to \Rr; \quad (y,z) \to  h(y,z)=\langle \Psi(y), \Psi(z)\rangle_\mathcal{H}.\vspace{-0.1cm}
$$
According to the Moore-Aronszajn theorem,  there is   a unique Hilbert space  of functions on $\Rr^p$, called  a reproducing kernel Hilbert space (RKHS)  \cite{steinwart2006explicit}, for which $h$ is a  symmetric positive definite kernel, also known as a reproducing kernel.  The advantage of such a construction is that the kernel trick\footnote{
Let us illustrate the kernel trick with the  polynomial kernel $ h(y,z)=(1+y^* z)^2$ and $p=2$. The kernel maps $\Rr^2$ onto the RKHS  $\mathcal{H}=\Rr^6$.
 Expanding the kernel, we obtain
$ 
 h(y,z)=(1+y^* z)^2=1+2y_1z_1+2y_2z_2+2y_1y_2z_1z_2+y_1^2z_1^2+y_2^2z_2^2,
 $
 where $y=\begin{pmatrix}y_1&y_2 \end{pmatrix}^*$ and $z=\begin{pmatrix}z_1 &z_2 \end{pmatrix}^*$ in $ \Rr^2$
so that we have 
 $h(y,z)=\Psi(y)^* \Psi(z)$,
 with $\Psi(y)=\begin{pmatrix}1& \sqrt{2}y_1&\sqrt{2}y_2&\sqrt{2}y_1y_2& y_1^2 &y_2^2  \end{pmatrix}^* \in  \Rr^6.$
Therefore, we can evaluate an inner product of vectors in  $\mathcal{H}$ by computing a function  of an inner product in $\Rr^2$.% and never evaluating explicitly $\Psi(y)$ and $\Psi(z)$.
}
%\footnote{
%Let us illustrate the kernel trick with the  polynomial kernel $ h(y,z)=(1+y^* z)^2$ and $p=2$. The kernel maps $\Rr^2$ onto the RKHS  $\mathcal{H}=\Rr^6$.
% Expanding the kernel, we obtain
%$ 
% h(y,z)=(1+y^* z)^2=1+2y_1z_1+2y_2z_2+2y_1y_2z_1z_2+y_1^2z_1^2+y_2^2z_2^2,
% $
% where $y=\begin{pmatrix}y_1&y_2 \end{pmatrix}^*$ and $z=\begin{pmatrix}z_1 &z_2 \end{pmatrix}^*$ in $ \Rr^2$
%so that we have 
% $h(y,z)=\Psi(y)^* \Psi(z)$,
% with $\Psi(y)=\begin{pmatrix}1& \sqrt{2}y_1&\sqrt{2}y_2&\sqrt{2}y_1y_2& y_1^2 &y_2^2  \end{pmatrix}^* \in  \Rr^6.$
%Therefore, we can evaluate an inner product of vectors in  $\mathcal{H}$ by computing a function  of an inner product in $\Rr^2$.% and never evaluating explicitly $\Psi(y)$ and $\Psi(z)$.
%} 
 can  be used to compute inner products in the RKHS $\mathcal{H}$ with a complexity equal to that required to  evaluate  the kernel $h$, which is in general independent of $\dim(\mathcal{H})$~\cite{bishop2006pattern}.   The key idea of their method called K-DMD is to use the kernel trick  to evaluate inner products with  eigen-vectors of $  \hat A^{\ell s}_k$.
 Assuming that the complexity for the evaluation of the kernel is $\mathcal{O}(p)$,   the overall complexity of the K-DMD algorithm is independent of $\dim(\mathcal{H})$ and ${\TT}$, \ie guaranteeing Constraint~\ref{property:3}. 
 %Interested readers may find a precise description of  Algorithm~\ref{algo:00} at the end of  this section.

 However, as proposed in  \cite{williams2014kernel},  K-DMD computes an approximation of reduced model \eqref{eq:koopman1} under restrictive assumptions.  In particular the  following assumptions are needed:  
{\it  i)}   the low-rank constraint in \eqref{eq:prob1} is ignored, \ie $A^\star_k=\hat A^{\ell s}_k$; %Therefore, state-of-the-art strategies implying two-stage approaches \cite{HeasHerzet17} will only provide  sub-optimal solutions.
{\it  ii)} the operator $\AAA$ is full-rank;
{\it  iii)} $\Psi^{-1}$ is linear;
{\it  iv)}   the  $ \Psi^{-1} \zeta_j$'s  belong to the span of $\YYY$. % as  the minimal energy solution of an under-determined approximation problem with $p$ unknowns and $m$ 
%Unfortunately, because $m$ is very small compared to $n$ in most cases, the second feature is a prominent limitation of the algorithm. %  it is unavoidable in the case of interest where $n \gg p$. 
%Indeed,  a necessary condition for optimality is that $n \le  m$. This configuraiton corresponds to an unlikely scenario where the number of observations is at least equal to the problem dimension. Moreover,  even though, $n \le  m$ would imply an intractable  complexity in $\mathcal{O}(m^3)$ for large $n$. %Therefore, this tractability implies that the kernel-based approach is useless and that a standard  EDMD solver is sufficient. 
%Furthermore,  the suitability of K-DMD  for  reduced modeling  is weak in the case $k < m$ and $p \gg 1$.  Indeed, we can show that the computation burden to compute the approximation is independent of the reduced model dimension $k$, but requires  $m^2$ vector operations  in $\Rr^p$. 
%More precisely,  
The reduced model computed by K-DMD  therefore does not meet Definition~\ref{property:1} because assumptions {\it i)} and {\it ii)} do not hold in general, nor Definition~\ref{property:2} because assumptions {\it iii)} and {\it iv)} do not hold either.   A   description of the algorithm can be found in Appendix~\ref{sec:kdmd}. 

On the other hand, while the kernel method~\cite{bouvrie2017kernel} is not in agreement  either with Definition~\ref{property:1}, it relaxes the assumptions {\it iii)} and {\it iv)} made by the K-DMD algorithm and  to some extent satisfies Definition~\ref{property:2}. Indeed,  the inverse map $\Psi^{-1}$ is learned by additional kernel regression steps in the sense of  (regularized) minimum distance estimation, similar  to \eqref{eq:defInverse2} to some extent.    The resulting reduced model is of the form \eqref{eq:model_koopman_approxObservable0}-\eqref{eq:model_koopman_approxObservable}, but still does not provide a diagonalised form like the targeted reduced  system~\eqref{eq:koopman1}. This implies that the kernel method~\cite{bouvrie2017kernel} is characterized by linear complexity in $T$, and therefore does not fulfill Constraint~\ref{property:3}. 
\vspace{-0.cm}
\vspace{-0.cm}%A consequence is that the lower $k$, the higher the ratio complexity/accuracy, which is obviously not a desirable property for reduced modeling.  

 \section{A Generalized Kernel-Based Algorithm} \label{sec:okDMD} \vspace{-0.cm} 
We present in this section our algorithm called  GK-DMD. The proposed method enables to computes the low-rank reduced model \eqref{eq:koopman1} for $\mathcal{H}$ being  a RKHS, with a complexity independent of $\dim(\mathcal{H})$ and $\TT$ and is relieved from the  assumptions made in K-DMD.  
 As for K-DMD, the GK-DMD exploits the kernel-trick and resorts to an analogous computation of eigen-functions. 
Its main innovation  in comparison to the latter state-of-the-art algorithm is that GK-DMD computes reduced model \eqref{eq:koopman1} based on the exact solution \eqref{eq:solAkOpt} of problem~\eqref{eq:prob1} and on the inverse definition~\eqref{eq:defInverse2}.   The GK-DMD algorithm, given in Algorithm~\ref{algo:3}  thus fulfills the the definitions and constraint  exposed in Section~\ref{sec:ReducedModelingProb}.   It will be presented in detail in Section~\ref{sec:algo}. In short, it consists of an off-line part, which computes a low-dimensional representation of $ A^\star_k$, and an on-line part, which uses the latter to evaluate an approximate trajectory for a given initial condition.

\subsection{Main results}  \label{sec:preuveOptimal} \vspace{-0.cm}

The proposed algorithm is based on the following two original results:

\begin{itemize}
\item  the right and left eigen-vectors of the optimal operator  $ A^\star_k$ belong to a low-dimensional sub-space of $\mathcal{H}$; their low-dimensional representations are tractable  and  computed  using the kernel function;  this  result is detailed in Section~\ref{sec:lowDimRep};  
\item  taking advantage of the fact that, in the reduced model~\eqref{eq:koopman1}, the inverse argument belongs to a low-dimensional subspace of $\mathcal{H}$, the high-dimensional minimization problem reduces to a tractable $p$-dimensional optimization problem; this result is detailed in the section~\ref{sec:kernelBasedInv}.
\end{itemize}

\subsubsection{Low-dimensional spectral representation of $ A^\star_k$}\vspace{-0.cm}\label{sec:lowDimRep}

Our algorithm relies on the following proposition. Its proof is detailed in Appendix \ref{app:3}. This result extends the closed-form eigen-decomposition  of operator $ \hat A^{\ell s}_m $ classically studied for EDMD (see \eg \cite{das2021reproducing,kawahara2016dynamic,klus2020eigendecompositions,korda2018convergence,williams2014kernel}), 
to the spectral characterization  of the low-rank operator $ A^\star_k $ at the heart of the low-rank version of EDMD of interest to us, which has not been studied in the literature. 
 Let $\{ \xi_i \}_{i=1}^k$  and $\{ \zeta_i \}_{i=1}^k$   denote the left and right eigen-vectors  of $ A^\star_k$ associated to its at most $k$ non-zero eigen-values $\{ \lambda_i \}_{i=1}^k$. \\
 
%\textcolor{red}{[
%\begin{itemize}
%\item Restructurer le resultat pour introduire mieux la portee de la proposition. En particulier dire que c est le resultat pour le probleme de KDMD etendu pour unecontrainte de bas rang, et que c est nouveau et different des resultats KDMD sans contraintes  proposes par WIlliam2015, Kawahara2016,Schwantes2015 ou Klus2019, 
%\item ca permet entre autre d eviter de regulariser ou de se projeter dans un sous espace arbitraire
%\end{itemize}]}
\vspace{-0.2cm} 

\begin{proposition}\label{prop:3.3} 
For $i=1,\ldots,k$, the left and right eigen-vectors  of $ A^\star_k$ and its eigen-values satisfy 
 \,\,$
%\begin{align*}
\xi_i=U_{\AAA} \tilde \xi_i \textrm{,}\quad \zeta_i=\hat P_k \tilde \zeta_i \quad \textrm{and}\quad \lambda_i=\tilde \lambda_{i},
%\end{align*}
$ where
 $\{ (\tilde \xi_i ,\tilde\lambda_{i}) \}_{i=1}^k$ and  $\{ (\tilde \zeta_i ,\tilde \lambda_{i}) \}_{i=1}^k$ denote  respectively  the first  $k$ right eigen-vectors and   eigen-values of  the  matrices in  $ \Rr^{m \times m}$\vspace{-0.1cm}
\begin{align}\label{eq:lowDimMatrices}
%(\tilde A_{\ell,k})^*= \Sigma_{\AAA}^\dagger V_{\AAA}^* P_0^* \BBB^* \AAA V_{\AAA} \Sigma_{\AAA}^\dagger \in \Rr^{m \times m},
R\,  \BBB^* \BBB\, S_k^*S_k \, \BBB^* \AAA\, R^*\quad \textrm{and} \quad  S_k\,  \BBB^* \BBB\, R^*\,R \, \AAA^* \BBB\, S_k^*,
\end{align}
with   $R= \Sigma_{\AAA}^\dagger V_{\AAA}^*$ and $ S_k= \textrm{diag}((\sigma^\ZZZ_{1})^{\dagger}\cdots (\sigma^\ZZZ_{k})^{\dagger}0\cdots 0) V_\ZZZ^* $. \vspace{-0.cm}\\%$P_0=V_{\AAA} \Sigma_{\AAA}^\dagger \Sigma_{\AAA} V_{\AAA}^* V_\ZZZ \textrm{rank}(\sigma^2_{\ZZZ,1}\cdots \sigma^2_{\ZZZ,k}0\cdots0) V_\ZZZ^*  V_{\AAA}\Sigma_{\AAA} \Sigma_{\AAA}^\dagger V_{\AAA}^* \BBB^* \BBB$
%\begin{align}
%%S&=  V_\ZZZ \textrm{diag}((\sigma^\dagger_{\ZZZ,1})^{2}\cdots (\sigma^{\dagger}_{\ZZZ,k})^{2}0\cdots 0) V_\ZZZ^* , \nonumber \\
%S.\nonumber
%\end{align}
\vspace{-0.1cm}
\end{proposition}
 
 \vspace{-0.cm}  This proposition gives  a closed-form decomposition for the $\xi_i$'s and the $\zeta_i$'s, the left and right eigen-vectors of  the optimal solution $ A^\star_k$ given in~\eqref{eq:solAkOpt} and supplies the related eigen-values $\lambda_i'$s.
The decompositions involve  operator  $\hat P_k$ defined  in \eqref{eq:Pk} and   operator $U_{\AAA}$ appearing in the  short-hand notation of the SVD of $\AAA$.
  %: 
%\begin{align}\label{eq:eigfuncApprox}
% \varphi_i(\theta)= \xi_i^* \Psi(\theta)=\tilde \xi_i^* R \AAA^* \Psi(\theta).
%\end{align}
%We mention that the condition $\tilde \zeta_i ^*  E  \tilde \xi_j=0,$ for $i \neq j$ is ensured as long as non-zero eigen-values are distinct.
A consequence of this proposition is that parameters  $\{(\xi_i,\zeta_i,\lambda_i)\}_{i=1}^k$  of  reduced model \eqref{eq:koopman1}  can be written in terms of their low-dimensional counterpart  $\{(\tilde \xi_i,\tilde \zeta_i,\tilde \lambda_{i})\}_{i=1}^k$. \vspace{-0.cm}%The latter 3-tuple will serve to design a tractable algorithm for the  computation of $\tilde x_{\TT}(\theta)$.  
% From now, we already notice that the 3-tuple enables to compute   $\varphi_i(\theta)$'s using \eqref{eq:eigfuncApprox0} and therefore coefficients $\nu_{i,t}$ in  \eqref{eq:koopman1}. \vspace{-0.4cm}%In the next section, we treat the remaining blocking computation of $\Psi^{-1} (\zeta_i \nu_{i,t})$.% for the computation of  $\AAA^* \AAA$, $\BBB^* \BBB$, $\BBB^* \AAA $ and $ \AAA^* \Psi(\theta)$  for any $\theta \in \Rr^p$. 
%More precisely, using the kernel trick to compute $\AAA^* \AAA$, $\BBB^* \BBB$, $\BBB^* \AAA $ and $ \AAA^* \Psi(\theta)$  for any $\theta \in \Rr^p$, the numbers of operations to evaluate and diagonalise  matrices $(\tilde A_{\ell,k})^*$ and $\tilde A_{r,k}$ scales respectively at most   in $m^2p$ and $m^3$. 
We notice that the first $k$  left  and right eigen-vectors of $ A^\star_k$ are normalized, \ie that  for $ i=1,\ldots,k$ the condition $\zeta_i^*\xi_i=1$ stands, if 
$
\tilde \zeta_i ^*  E  \tilde \xi_i=1$ with $
E=S_k\BBB^* \AAA R^*$.  
This normalization condition can be verified using some  algebraic calculus. 
We ensure condition $\tilde \zeta_i ^*  E  \tilde \xi_i=1$  for $ i=1,\ldots,k$   by first computing  the complex number $\gamma_i=\tilde \zeta_i ^*  E  \tilde \xi_i$  and then  rescaling the eigen-vector $\tilde \zeta_i$  by $\gamma_i^{-1}$.

\subsubsection{Kernel-based inversion for reduced modeling}\vspace{-0.cm}\label{sec:kernelBasedInv}
 The low-dimensional representation of eigen-vectors of $A^\star_k$  provided in Proposition~\ref{prop:3.3}  constitutes an essential ingredient of the GK-DMD algorithm. However, to achieve the design of this algorithm, it  remains to provide a feasible manner to evaluate the inverse map  $\Psi^{-1}$ in \eqref{eq:koopman1}, with the inverse map  definition in \eqref{eq:defInverse2}. Once more, the  idea  consists in relying  on the kernel trick in order to evaluate the inverse with a complexity independent of $\dim(\mathcal{H})$. 
 The following result shows that the sought reduced model boils down to solving a $p$-dimensional minimization problem involving scalar product in $\mathcal{H}$ evaluated thanks to the kernel-trick.\\
 
\begin{proposition}\label{prop:inverse} 
\vspace{-0.cm}
For a symmetric positive definite kernel  $h$, using the low-rank operator Definition \ref{property:1} and the inverse map Definition \ref{property:2}, the reduced model~\eqref{eq:koopman1}   of the high-dimensional dynamics~\eqref{eq:model_init}   solves  at any point  $\theta \in \Rr^p$ 
\begin{align}\label{eq:defInverse3}
\tilde x_{\TT}(\theta)&\in  \arg\min_{z\in \Rr^p} \left(  h(z,z)-2\sum_{i=1}^m  g_i^{\theta,{\TT}} {h(y_i,z)} \right),
\end{align}
{where} $g^{\theta,{\TT}}=S_k^* (\tilde \zeta_1 \cdots \tilde \zeta_k)  \begin{pmatrix}\tilde \lambda^{{\TT}-1}_{\ell,1} \varphi_1(\theta) &\cdots& \tilde \lambda^{{\TT}-1}_{\ell,k}\varphi_k(\theta)\end{pmatrix}^* \in  \Rr^{m}$ and where for  $i=1,\ldots, k$    the $i$th   eigen-function  is given by
 \begin{align}\label{eq:eigfuncApprox0}
 \varphi_i(\theta)=  \tilde \xi_i^* R \AAA^*  \Psi(\theta).\vspace{-0.4cm}
\end{align}
\end{proposition}

The proof is provided  in Appendix~\ref{app:1.2}. Sufficient conditions for the existence of minimizer \eqref{eq:defInverse3} are provided and discussed for various kernel choices  in Appendix~\ref{app:1.3}.
  A minimizer \eqref{eq:defInverse3} can be computed (up to some accuracy) using standard  optimization methods with a complexity independent  of $\dim(\mathcal{H})$ and of $T$, thus satisfying Constraint \ref{property:3}. Moreover, the gradient of the objective is in general  closed-form, which enables the use of efficient  large-scale optimization techniques such as limited memory quasi-Newton methods~\cite{nocedal2000numerical}. In this case, the complexity to compute the inverse is  linear in $p$.

 \vspace{-0.cm} \subsection{The GK-DMD algorithm} \label{sec:algo} \vspace{-0.cm} 
\begin{algorithm}[h]
 \begin{algorithmic}[0]
\State $\bullet$ \textbf{Off-line.}\\
\hspace{0.3cm}{\bf Inputs:}   $x_t(\thetaLearn_i)$'s 
\begin{algorithmic}[0]
%\State {\bf Inputs:}  the $x_t(\thetaLearn_i)$'s 
\State 1) Compute matrices   $\AAA^*  \AAA$, $\BBB^*  \BBB$, $\BBB^*  \AAA  $  in $\Rr^{m \times m}$  with the kernel trick.%=U_{\hat\Q}\Sigma_{\hat\Q} V_{\hat\Q}^* $.%\ie and  the SVD of matrix $\AAA$ %, we obtain are able to compute $\mathbf{w}\in \Rr^{m \times m}$ satisfying 
%\begin{align*}
%\AAA&= U_a \,\begin{pmatrix}\textrm{diag}({\sigma_a}^{\frac{1}{2}})\\ 0_{n-\ell}\end{pmatrix}    V_a ^* .
%\end{align*}
%\State 2)  Define matrix   $D_k=U_{\hat\Q}^* \hat \R_k V_{\hat\Q}\Sigma_{\hat\Q} \in \Rr^{m\times m}$.
\State 2) Compute $(V_{\AAA} , \Sigma_{\AAA})$ by eigen-decomposition of  $\AAA^*   \AAA$. 
%
%%\State 1)  Form matrix $\AAA$ and $\BBB$ as defined in \eqref{eq:matrixAB}. %, we obtain are able to compute $\mathbf{w}\in \Rr^{m \times m}$ satisfying 
%\State 1) Compute step 1) and 2) of Algorithm~\ref{algo:00}  using the kernel trick~\cite{bishop2006pattern}.$^3$
%%the matrices $\AAA^* \AAA$, $\BBB^* \BBB$ and $\BBB^* \AAA  $ using the kernel trick with the data  $\XXX$ and $\YYY$.%=U_{\hat\Q}\Sigma_{\hat\Q} V_{\hat\Q}^* $.%\ie and  the SVD of matrix $\AAA$ %, we obtain are able to compute $\mathbf{w}\in \Rr^{m \times m}$ satisfying 
%%%\begin{align*}
%%%\AAA&= U_a \,\begin{pmatrix}\textrm{diag}({\sigma_a}^{\frac{1}{2}})\\ 0_{n-\ell}\end{pmatrix}    V_a ^* .
%%%\end{align*}
%%%\State 2)  Define matrix   $D_k=U_{\hat\Q}^* \hat \R_k V_{\hat\Q}\Sigma_{\hat\Q} \in \Rr^{m\times m}$.
%%\State 2) Compute $(V_{\AAA} , \Sigma_{\AAA})$ through the eigen-decomposition of  $\AAA^* \AAA$. 
%%%\State 4) Compute the projector $ \mathbb{P}_{\AAA^*}= %\AAA^\dagger\AAA=
%%%V_{\AAA}\Sigma_{\AAA}\Sigma_{\AAA}^\dagger V_{\AAA}^*$.  
%%%\State 3) The $i$th DMD mode corresponding to the eigen-value $\lambda_i$ is then given by 
%%%$$ \phi_i=.$$
%%\State 3) Compute $\ZZZ^* \ZZZ$ where $\ZZZ$ is given by \eqref{eq:Z}  and given matrix $\BBB^* \BBB$.% and projector $ \mathbb{P}_{\AAA^*}= V_{\AAA}\Sigma_{\AAA}\Sigma_{\AAA}^\dagger V_{\AAA}^*$.
\State 3) Compute $(V_{\ZZZ} , \Sigma_{\ZZZ})$ by eigen-decomposition of  $\ZZZ^* \ZZZ$ with  $\ZZZ$ given by \eqref{eq:Z}. 
\State 4) Compute the two matrices given  in  Proposition~\ref{prop:3.3}  and compute  their eigen-vector/eigen-value couples $\{ (\tilde \xi_i ,\tilde \lambda_{i}) \}_{i=1}^k$ and $\{ (\tilde \zeta_i ,\tilde \lambda_{i}) \}_{i=1}^k$.  
\State 5)  Rescale each right eigenvector $\tilde \zeta_i$ by factor $(\tilde \zeta_i^*\, E\, \tilde \xi_i)^{-1}$ with
$
E$=$S_k\BBB^* \AAA R^*.$\\
 \textbf{Outputs}: $R$, $S_k$, $ \tilde \xi_i$'s, $ \tilde \zeta_i$'s and $\tilde \lambda_i$'s
\end{algorithmic}
\end{algorithmic}
\begin{algorithmic}[0]
\State $\bullet$ \textbf{On-line.}\\
\hspace{0.3cm} \textbf{Inputs}:  off-line outputs and $\theta$
\begin{algorithmic}[0]
\State 6) Compute  $ \AAA^*  \Psi(\theta)$ in $\Rr^{m}$ with the kernel trick.
\State 7) Compute  eigen-functions  $\{ \varphi_i(\theta)\}_{i=1}^k$ defined in \eqref{eq:eigfuncApprox0}. \vspace{-0.cm}
\State 8) Compute  $ \tilde x_{\TT}(\theta) $ solving \eqref{eq:defInverse3}; \\% or \eqref{eq:invers3}. 
 \textbf{Output}: $\tilde x_{\TT}(\theta)$.%, $\delta$ \remCH{changer tolerence par "stopping criterion"}
\end{algorithmic}
\end{algorithmic}
\caption{: GK-DMD   \label{algo:3}} 
\end{algorithm}
 \vspace{-0.cm}

We begin by detailing the off-line computation.  According to Proposition~\ref{prop:3.3}, the low-dimensional representation of the right and left eigen-vectors, and the associated  eigen-values of the optimal operator  $ A^\star_k $  are obtained by eigen-decomposition of matrices given in \eqref{eq:lowDimMatrices}.  As detailed hereafter, the  first five steps  of the algorithm precisely compute  this low-dimensional representation. Relying on the kernel function, step 1 first computes the inner products in $\mathcal{H}$ needed to build  matrices 
$\AAA^*  \AAA$, $\BBB^*  \BBB$, $\BBB^*  \AAA  $ in $\Rr^{m\times m}$. In steps~2 and 3, the matrices $R$ and $S_k$ in $\Rr^{m\times m}$ are  obtained by  eigen-decomposition of  $\AAA^*   \AAA $  and $\ZZZ^* \ZZZ$ belonging also to $\Rr^{m\times m}$. These three first steps enable to obtain matrices given in \eqref{eq:lowDimMatrices} by  simple matrix products. Step 4 of the algorithm  computes the right and left eigen-vectors and the associated eigen-values of matrices given in~\eqref{eq:lowDimMatrices}. Step~5 finally normalizes the latter eigen-vectors, as detailed in Section~\ref{sec:lowDimRep}.

In the on-line part of the algorithm, the low-dimensional representation given by the outputs of the off-line part is exploited to compute the reduced model output $\tilde x_{\TT}(\theta)$, for a given initial condition $\theta$. As discussed in Section~\ref{sec:kernelBasedInv}, the computation of $\tilde x_{\TT}(\theta)$ is obtained by solving the $p$-dimensional optimization problem \eqref{eq:defInverse3}. The objective function of this problem depends on kernel functions weighted by the $m$-dimensional coefficient vector $g^{\theta,{\TT}}$ defined in Proposition~\ref{prop:inverse}. This  vector  depends itself on the low-dimensional representation of  operator  $ A^\star_k$ computed in the off-line part and on the eigen-functions  $\{ \varphi_i(\theta)\}_{i=1}^k$. These eigen-functions  are obtained in steps 6 and 7 of the algorithm : step 6 uses  the kernel trick to compute the $m$-dimensional vector $ \AAA^*  \Psi(\theta)$  while  step 7 deduces the eigen-functions  \eqref{eq:eigfuncApprox0} using the low-dimensional representation. With these eigen-functions, we  then compute the vector component $g^{\theta,{\TT}}_i$ weighting the kernel function $h(y_i,z)$ in the objective function of  \eqref{eq:defInverse3}. Step 8 finally solves this optimization problem and provides  the reduced model    $\tilde x_{\TT}(\theta)$.

 \vspace{-0.cm} \subsection{Performance analysis}  \vspace{-0.cm} 
 In this section, we begin by analyzing the advantage of the GK-DMD algorithm in terms of computational complexity and then show that the approximation error is optimal  in some sense in the RKHS. Moreover, we show that  the norm of the learning approximation error is tractable without any increase of the algorithm complexity.
\subsubsection{Complexity} 
% In particular, it
%%requires the computation of an SVD of a matrix belonging to  $\Rr^{n \times m }$, matrix multiplications 
%involves $m^2$ vector products in $\Rr^p$, $m^2$ vector products in  $\Rr^m$ and eigen-decompositions and inversion of matrices belonging to  $\Rr^{m \times m}$. 
Assuming a complexity  $\mathcal{O}(p)$ for the evaluation of the kernel function\footnote{This property  is true for most standard kernels, \eg polynomial or Gaussian kernels~\cite{bishop2006pattern}.},  the overall complexity of the proposed algorithm scales in  $\mathcal{O}(m^3+m^2p)$, just as for  K-DMD, as detailed in Appendix~\ref{sec:kdmd}. 
 More precisely, in the off-line part,  eigen-decompositions (involving step 2, 3 and 4) are computed\footnote{In the case of large $m$, randomized algorithms for low-rank matrix approximation can significantly reduce the cubic complexity of the proposed kernel method~\cite{bach2005predictive,tropp2023randomized,yang2012nystrom}.} in $\mathcal{O}(m^3)$, while the kernel evaluations (step 1) and the rescaling (step 5) are respectively done in $\mathcal{O}(m^2p)$ and  $\mathcal{O}(m^2k)$, with $k\le m \le p$.    We remark that those off-line  complexities are independent of ${\TT}$ thanks to the diagonalization of $A^\star_k$,  and independent of $\dim(\mathcal{H})$ due to the use of the kernel-trick  in the first and last steps of the algorithm. 
 
%A  closer inspection shows that the computational burdens of K-DMD and GK-DMD are both dominated by an off-line complexity  in $\mathcal{O}(m^2(m+p))$, because of the matrix product computation in step 1) and the eigen-decompositions in step 2), 3) and 4).
Reduced modeling is very concerned by the on-line computational cost, \ie complexity of computation steps depending on the input $\theta$.  As   typically $k\ll p$, GK-DMD is attractive  by its   on-line complexity  in $\mathcal{O}(m^2k+mp )$, thus it scales linearly with respect to the dimension of the reduced model $k$ or the ambient dimension $p$,  in comparison to   $\mathcal{O}(m^2p)$ operations for K-DMD.
 Indeed, the matrix-vector product  $ \AAA^*  \Psi(\theta)$  (step 6)  and  the inversion (step 8) are both computed in $\mathcal{O}(pm)$ operations using standard optimization tools, while eigen-functions  (step 7) require  $\mathcal{O}(m^2k)$  operations. 

  Note that, as well as being constant in $T$ and  $\dim(\mathcal{H})$, the complexity characterizing the  reduced dynamics in GK-DMD is of complexity  $\mathcal{O}(m^2k)$. It is thus also constant with respect to the dimension of the ambient space $p$, which is a standard desired feature for reduced models. Note that the initialization or the final reconstruction step of any reduced model will involve  at least $p$ operations. In the case of GK-DMD, the mapping from the initial state (living in the ambient space of dimension $p$) to the reduced space (in step 6) or the  inverse mapping from the reduced space to the ambient space (in step 8) are computed in  $\mathcal{O}(mp)$.
 
 % Nevertheless, as the complexity is  quadratic in $m=N(T-1)$, the GK-DMD method is particularly well suited for learning setups where $N$ and $T$ are not too large. 
  \vspace{-0.cm}
 \subsubsection{Learning accuracy} 
 The  learning error  of our GK-DMD algorithm is optimal in the sense it is equal to the minimal  Hilbert-Schmidt norm  of the error achievable in the RKHS for the minimization problem~\eqref{eq:prob1}. The norm of this optimal error, given by a generalization of \cite[Theorem 4.1]{HeasHerzet17}   to  separable infinite-dimensional Hilbert spaces, is closed-form\vspace{-0.2cm}
 \begin{align}\label{eq:optError}
 \|\BBB -A_k^\star  \AAA \|^2_{\mathcal{HS}}=  \sum_{i=k+1}^{m} (\sigma_i^\ZZZ)^{2}   +   \sum_{i=i^\star}^m   \sum_{j=1}^m ( \sigma_j^{\BBB} )^2\left(( v_j^{\BBB})^* v_i^{\AAA}\right)^2 ,\vspace{-0.2cm}
 \end{align}
with $i^\star=\rank(\AAA)+1$~\cite{HeasHerzet18Maps}.

Moreover, it is worth noticing that the optimal error norm~\eqref{eq:optError} is tractable. It  possibly involves a little extra computation which does not increase the complexity of Algorithm~\ref{algo:3}. More precisely, we notice that the left singular vectors $\{v_i^{\AAA}\}_{i=1}^m$ and $\{v_i^{\ZZZ}\}_{i=1}^m$ respectively of ${\AAA}$ and ${\ZZZ}$ and their associated singular values $\{\sigma_i^{\AAA}\}_{i=1}^m$ and $\{\sigma_i^\ZZZ\}_{i=1}^m$ have been precomputed in step 2) and 3) of the algorithm. We distinguish the two following cases depending on the value of $i^\star$, which  equals to the number of non-zero singular values in the set $\{\sigma_i^{\AAA}\}_{i=1}^m$ plus one. 
\begin{itemize}
\item If $i^\star=m+1$, \ie $\AAA$ is full rank, the second term on the right-hand side vanishes, and the error norm is directly computable. 
\item Otherwise, in order to evaluate the error norm, we need  to make the extra computation of  the left singular vectors
 $\{v_i^{\BBB}\}_{i=1}^m$ and singular values $\{\sigma_i^{\BBB}\}_{i=1}^m$  of $\BBB$. They are obtained  by eigen-decomposition of matrix $\BBB^*\BBB\in\Rr^{m \times m}$, precomputed using the kernel trick in step 1) of the algorithm. This additional eigen-decomposition will involve a complexity in $\mathcal{O}(m^3)$, \ie preserves the global algorithm complexity  in  $\mathcal{O}(m^2(m+p))$.
 \end{itemize}
\section{Numerical Simulations}\label{sec:numEval} \vspace{-0.cm}

In this section, we assess four data-driven reduced modeling methods for  the approximation of two high-dimensional systems. 

  \subsection{Experimental setup}

Our benchmark consists of the four following  algorithms:
\begin{itemize}
\item low-rank  DMD (LR-DMD) \cite[Algorithm 3]{HeasHerzet17},
\item  total-least-square  DMD (TLS-DMD) \cite{hemati2017biasing}, 
\item  kernel-based DMD (K-DMD) \cite{williams2014kernel}, 
\item the proposed generalized kernel DMD (GK-DMD), \ie Algorithm \ref{algo:3}.
\end{itemize}
For the K-DMD and GK-DMD algorithms,   we use a quadratic   polynomial kernel  or  a Gaussian kernel with a standard deviation of $10$~\cite{bishop2006pattern}. Details on these kernels are provided in Appendix \ref{sec:examples}.  The table below summarizes   the off-line and on-line complexities of the different algorithms. 
 \begin{table}[h!]
  \begin{center}\vspace{-0.35cm}
\begin{tabular}{c|c|c|c|c}
                   & { LR-DMD} & { TLS-DMD} & { K-DMD} & { GK-DMD} \\
                   \hline 
off-line  & {\footnotesize $\mathcal{O}(m^2(m+p))$ }& {\footnotesize $\mathcal{O}(m^2(m+p))$ } & {\footnotesize $\mathcal{O}(m^2(m+p))$} & {\footnotesize $\mathcal{O}(m^2(m+p))$} \\
on-line   & {\footnotesize $\mathcal{O}(pk )$ }& {\footnotesize $\mathcal{O}(\TT pm)$ } & {\footnotesize $\mathcal{O}(pm^2)$ }& {\footnotesize $\mathcal{O}(pm +m^2k)$}\\
\end{tabular}\label{tab:1}
   ~\\   ~\\
\caption{\small  Off-line and on-line complexities of reduced modeling algorithms.  We recall that $p$ is the ambient state dimension, $k$ is the reduced model dimension, $\TT$ is the predicted trajectory length and $m$ is the size of the training data set.}\vspace{-0.6cm}\label{Tab:1}
\end{center}
\end{table}
     
These  algorithms are used to approximate  two  high-dimensional systems. 
    
\begin{itemize}
\item A Rayleigh-B\'enard convection \cite{chandrasekhar2013hydrodynamic} model, which is a standard benchmark   in meteorology  for the evolution of  vorticity and temperature fields. Convection  is driven by  two coupled non-linear partial differential equations.  

\item A divergence-free fractional Brownian motion (fBm) evolution model, which is inspired by the turbulence phenomenology. The aim is to assess the ability of reduced modeling methods to capture the non-linear dynamics and the precise fractal structure of eddies. %The initial states  of dimension $p=512$   are  bi-dimensional divergence-free fBm vector fields  discretized on a square grid $16 \times 16$. Vector field evolution is driven by a discrete system according to a diffusive quadratic  model.
\end{itemize}
A precise description of these  high-dimensional systems,  together with details on the experimental  setting used in our numerical simulations, are provided below. \medskip

{\bf Rayleigh-B\'enard convection.}\,\,
Rayleigh-B\'enard convection \cite{chandrasekhar2013hydrodynamic}  is a standard benchmark model  in meteorology. Convection is driven by  two coupled  partial differential equations. 
    Let $\nabla=(\partial_{s_1},\partial_{s_2})^* $, $\nabla^{\perp}=(\partial_{s_2},-\partial_{s_1})^*$ and $\Delta=\partial^2_{s_1}+\partial^2_{s_2}$ denote the gradient, the curl and the Laplacian with respect to the two spatial dimensions $(s_1,s_2)$. 
     Boundary conditions are $1$-periodic  along $s_1$ and of Dirichlet type\footnote{In order to simplify the Fourier-based numerical implementation of the model, we will assume   periodicity for the discretised system in the two  spatial  directions.}  on $s_2$.   At any  point of the unit cell $\s=(s_1,s_2)\in [0,1]^2$  and for any time  $t \ge 1$, the temperature  $\tau(\s,t)\in \Rr $,  the vorticity  $b(\s,t)\in \Rr$ and  the velocity ${v}(\s,t) \in \Rr^2 $ in the cell satisfy \vspace{-0.3cm}
\begin{align*}
 \left\{\begin{aligned}
&\partial_t b(\s,t) + {v}(\s,t)^* \nabla b(\s,t)-\operatorname{Pr}\Delta b(\s,t) - \operatorname{Pr} \operatorname{Ra} \partial_{s_1} \tau (\s,t)=0,\\
 &\partial_t  \tau(\s,t) + {v}(\s,t)^* \nabla \tau(\s,t)- \Delta \tau(\s,t) -\partial_{s_1} \Delta^{-1} b(\s,t)=0,
\end{aligned}\right. 
\end{align*} 
where velocity is related to  vorticity  according to
${v}(\s,t)= \nabla^{\perp}\Delta^{-1} b(\s,t),$  
and where  $\Delta^{-1}$ represents the inverse of $\Delta$ defined in the Fourier domain. We set the Rayleigh  number to $ \operatorname{Ra}=1.5e5$ and the Prandtl number to $\operatorname{Pr}=1.0e1$. We use a Runge-Kutta fourth-order time discretization. For spatial discretization, we use  a Fourier-based implementation of spatial derivatives except for the advection term for which we use a second-order finite difference  scheme.  For time-integration of the dynamic model, we
 use a time step of $1e-4$.
 We obtain  a discrete system of the form of~\eqref{eq:model_init} with $x_t=({b}_t, {\tau}_t)^*\in \Rr^{p}$, and $p=4096$, where ${b}_t$'s and ${\tau}_t$'s are  spatial discretizations on a grid of size $64 \times 32 $ of vorticity and temperature fields at time~$t$. In our experiments, we assume that the  initial condition belongs to the Lorenz attractor~\cite{Lorenz63}. The initial state is of  the form
$b(\s,1)=\kappa_b\sin(a_b s_1)\sin(\pi s_2),$ $\tau(\s,1)=\kappa_{\tau_1}\cos(a_\tau s_1)\sin(\pi s_2)-\kappa_{\tau_2}\sin(2\pi s_2),$
with the parameter vector  $(a_b, a_\tau, \kappa_b, \kappa_{\tau_1},\kappa_{\tau_2}) \in \Rr^5$. 
In our experiments, we sample  parameters to generate the training and test data sets using a uniform distribution on the
hyper-cube  $(a_b, a_\tau, \kappa_b, \kappa_{\tau_1},\kappa_{\tau_2}) \in[0, 0.1]^5$.\medskip
% for $t=1,\ldots,10$ (resulting in 100 states $x_t(\thetaLearn_j)$). Examples of $x_t(\thetaLearn_j)'s$ are displayed in Figure~\ref{fig:1}.
% The test data set is generated as follows:   10 initial conditions  are set as  $\theta_j=x_{10} (\thetaLearn_j)$  and  trajectories
%$x_t(\theta_j)$  for $t=1,\ldots,10$ and $j=1, \ldots,10$ are computed in the same way as for
%the training data set.

%---
%For training the reduced model, we consider  the states of the high-dimensional system \eqref{eq:RB} for  $t=1,\ldots,T=10$, using an integration time step of $1e-4$.  
%We draw $N=10$ identical and independently distributed  samples of the initial condition parameter vector  using a uniform distribution on the hyper-cube  $[0, 0.1]^5$. 
%In consequence, we obtain $m=N  (T-1)=90$ data pairs $(x_t(\thetaLearn_j),x_{t+1}(\thetaLearn_j))$  for time index $t=1,\ldots, T$ associated to the $N$  initial conditions
%  $\{\thetaLearn_j\}_{j=1}^{N}$.  Examples of snapshots $x_t(\thetaLearn_j)$ are presented in Figure~\ref{fig:1}.
%Then, we  set $Ntheta=10$ new initial conditions $\{\theta_j\}_{j=1}^{\NN}$ as $\theta_j=x_{T} (\thetaLearn_j)$ and compute  trajectory approximations   $\tilde{x}_t(\theta_j)$ for   $t=1,\ldots, \TT=10$ using  the learned reduced models. We mention that the $\thetaLearn_j$'s are rescaled so that there is a good agreement with the condition $\|y_i\|_2 \ll 1$ required by approximations~\eqref{eq:invers1} and \eqref{eq:invers2}

{\bf Divergence-free fractional Brownian motion evolution.}\,\,
A simple model inspired by the turbulence phenomenology  is  proposed to evaluate the capability of reduced modeling methods.  The states  are discrete motion fields defined on a square grid satisfying 
 \eqref{eq:model_init}  with the quadratic model  $$f(x_t)=(x_{t-1}+1)^2 +\alpha\, L\, x_{t-1}-1,$$
	  where  the square power is taken component-wise,  the diffusion coefficient is $\alpha=0.5$ and  $L\in \Rr^{p\times p}$ is a second-order finite difference  approximation of  the (bi-variate) two-dimensional Laplacian operator. 
	Initial states used to generate the training and test data sets are samples from a  bi-dimensional divergence-free fractional Brownian motion (fBm) field of Hurst exponent $1/3$. The ambient dimension is set  to $p=512$ and the size of the square grid is set $16 \times 16$. The fBm vector fields are parametrized  according to the fractional wavelet representation proposed in~\cite[Proposition 3.1]{heas2014self}, using 18  random coefficients drawn according to a normal law. \vspace{-0.cm}\medskip

Note that the complexity of evaluating a trajectory using these high-dimensional systems is generally of the order of $\mathcal{O}(pT(\delta t)^{-1})$, where $\delta t$ represents the unit time discretiszation step. To obtain a stable simulation, $\delta t$ often has to be very small, which makes trajectory evaluation very demanding. The on-line complexity $\mathcal{O}(pm +m^2k)$ offered by  GK-DMD for approximating a trajectory is very attractive in this context. For example,  in our Rayleigh-B\'enard convection framework ($p$=$4096$, $m$=$90$, $T$=$10$, $\delta t=1$e-$4$), the complexity gain provided by GK-DMD is for any $k\le m$  of  more than two orders of magnitude.\\

  \subsection{Evaluation criteria}

     We study the evolution with respect to the rank  $k$ of   the
     \begin{itemize}  
  \item {\bf  reconstruction error}    \vspace{-0.3cm}
 $$
\epsilon_{rec}= \left(\sum_{j=1}^{\NN} \sum_{t=1}^{\TT-1}\frac{ \|\tilde x_2( x_{t}(\theta_j))-x_{t+1}(\theta_j)\|^2_2}{ \|x_{t+1}(\theta_j)\|^2_2}\right)^{1/2},
$$
   for a set of initial conditions $\{\theta_j\}_{j=1}^{\NN}$ where $\NN$ is given (which differ from the set of initial conditions $\{\thetaLearn_j\}_{j=1}^{\NNPrim}$ used for training); it measures the overall discrepancy
 between the true state $x_{t+1}(\theta_j)$ at time $t+1$ and the approximated state
 $\tilde x_2(x_{t}(\theta_j))$ predicted with the reduced model from the true state  at time $t$ for $t=1,\ldots,T-1$;
 \item  {\bf  learning error} \vspace{-0.3cm}
 $$
\epsilon_{\text{learn}}=\left(\sum_{j=1}^{\NNPrim} \sum_{t=1}^{\TTPrim-1}\frac{ \|\tilde x_2( x_{t}(\thetaLearn_j))-x_{t+1}(\thetaLearn_j)\|^2_2}{ \|x_{t+1}(\thetaLearn_j)\|^2_2}\right)^{1/2};
$$
it measures the overall discrepancy between the  state $x_{t+1}(\thetaLearn_j)$  and the prediction $\tilde x_2( x_{t}(\thetaLearn_j))$ at time $t$ for $t=1,\ldots, \TTPrim-1$; it reveals the capability of the reduced model to  approximate the training data used during the learning stage.
     \end{itemize}  
We mention that a small learning error $\epsilon_{\text{learn}}$ will not necessarily yield a small reconstruction error~$\epsilon_{rec}$. A small   $\epsilon_{\text{learn}}$ may  be related to  model overfitting. The learning error will nonetheless help to understand the behavior of the different methods.

% Another quantity of interest is the {\bf learning error} defined as
% $$
%\epsilon_{\text{learn}}=\left(\sum_{j=1}^{N} \sum_{t=1}^{T-1}\frac{ \|\tilde x_2( x_{t}(\thetaLearn_j))-x_{t+1}(\thetaLearn_j)\|^2_2}{ \|x_{t+1}(\thetaLearn_j)\|^2_2}\right)^{1/2},
%$$
%Measuring the discrepancy between the  state $x_{t+1}(\thetaLearn_j)$  and the prediction $\tilde x_2( x_{t}(\thetaLearn_j))$ reveals the capability of the reduced model to  reproduce the snapshots used during the learning phase.
%We mention that a small learning error $\epsilon_{\text{learn}}$ will not necessarily yield a small reconstruction error~$\epsilon_{rec}$. A small value for $\epsilon_{\text{learn}}$ may  be related to  model overfitting. Nonetheless,  this error helps to understand the behavior of the different methods.

%In the next experiments, we assess these two errors with the four methods
%using $\NN=N=10$ and $\TT=T=20$ on two different dynamic model that
%will be described in the next section.

  \begin{figure}[!t]
\centering
\vspace{-0.cm}\begin{tabular}{cccc}
\multicolumn{3}{c}{Rayleigh-B\'enard convection}\\
\hline \\
\hspace{-0.5cm} \includegraphics[height=0.1\columnwidth]{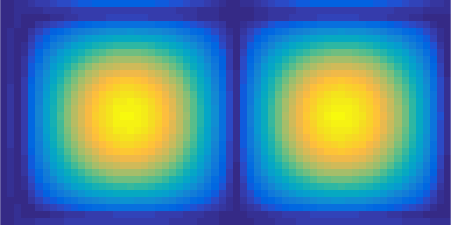}  &\hspace{-2.cm}  \includegraphics[height=0.1\columnwidth]{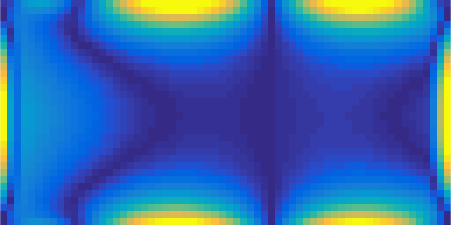}  &\hspace{-3.5cm}  \includegraphics[height=0.1\columnwidth]{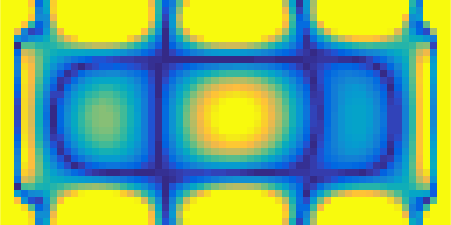} &\hspace{-2.5cm} \vspace{-0.05cm}\footnotesize{Absolute vorticity}\\
&&&\hspace{-2.5cm}\includegraphics[width=0.175\columnwidth]{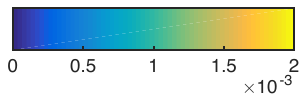} \\  
\hspace{-0.5cm}  \includegraphics[height=0.1\columnwidth]{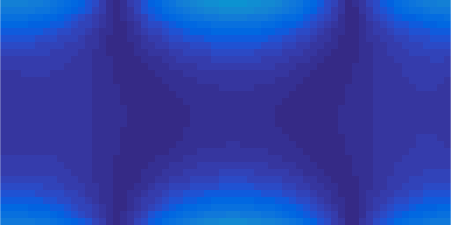}  &\hspace{-2cm}  \includegraphics[height=0.1\columnwidth]{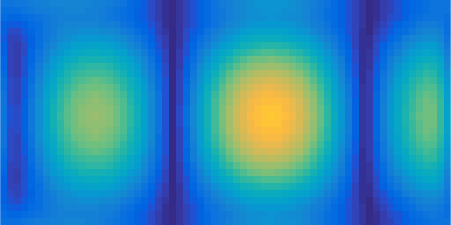}  &\hspace{-3.5cm}  \includegraphics[height=0.1\columnwidth]{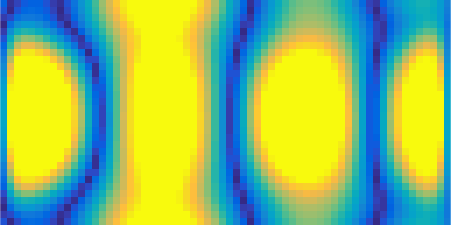}&\hspace{-2.5cm}\vspace{-0.05cm}  \footnotesize{Temperature}\\
 &&&\hspace{-2.5cm}\includegraphics[width=0.175\columnwidth]{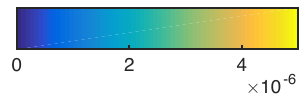} \vspace{-0.3cm} \\
\multicolumn{3}{c}{Divergence-free fBm evolution model}\\
\hline \\
\hspace{-0.5cm} \includegraphics[height=0.3\columnwidth]{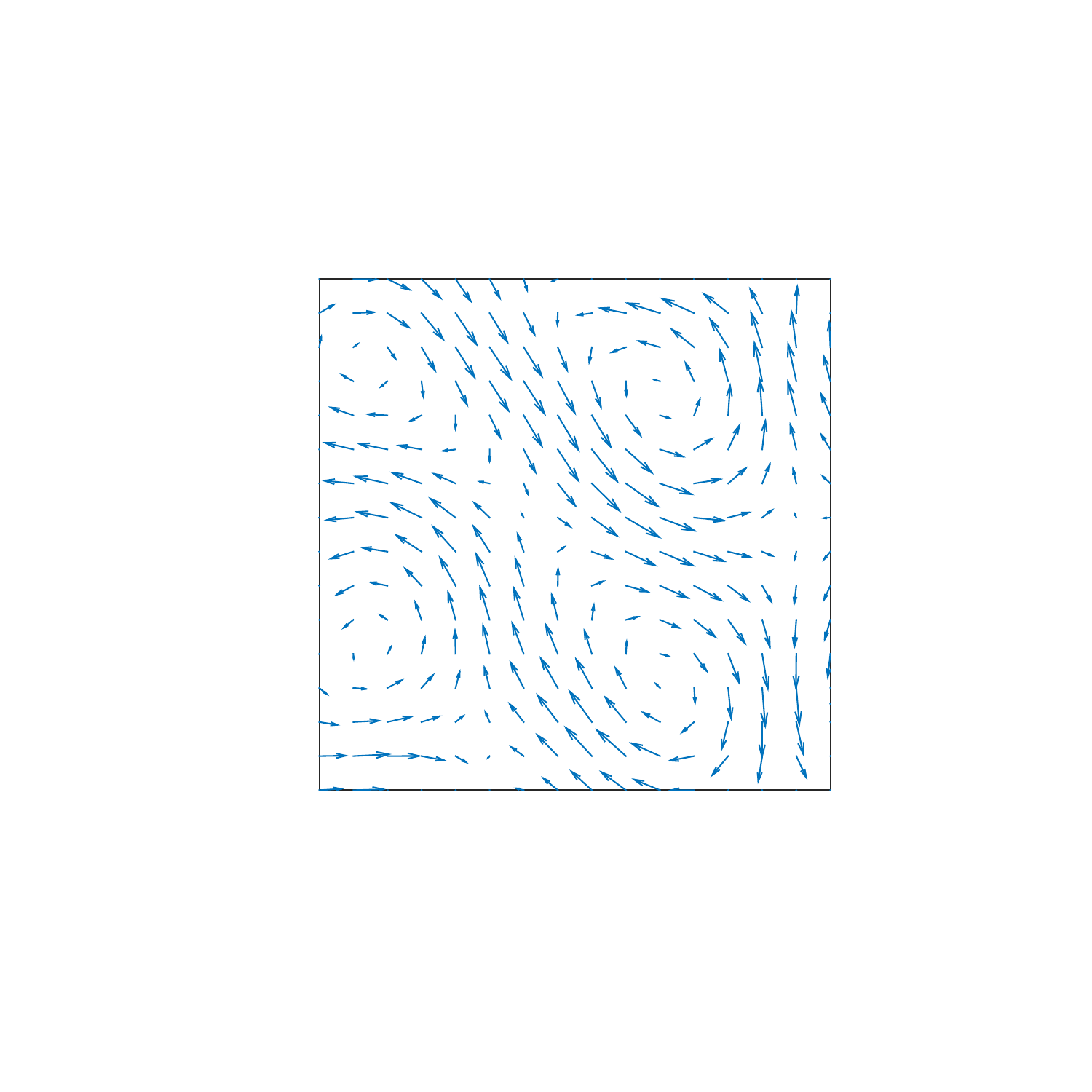}  &\hspace{-0.4cm}  \includegraphics[height=0.3\columnwidth]{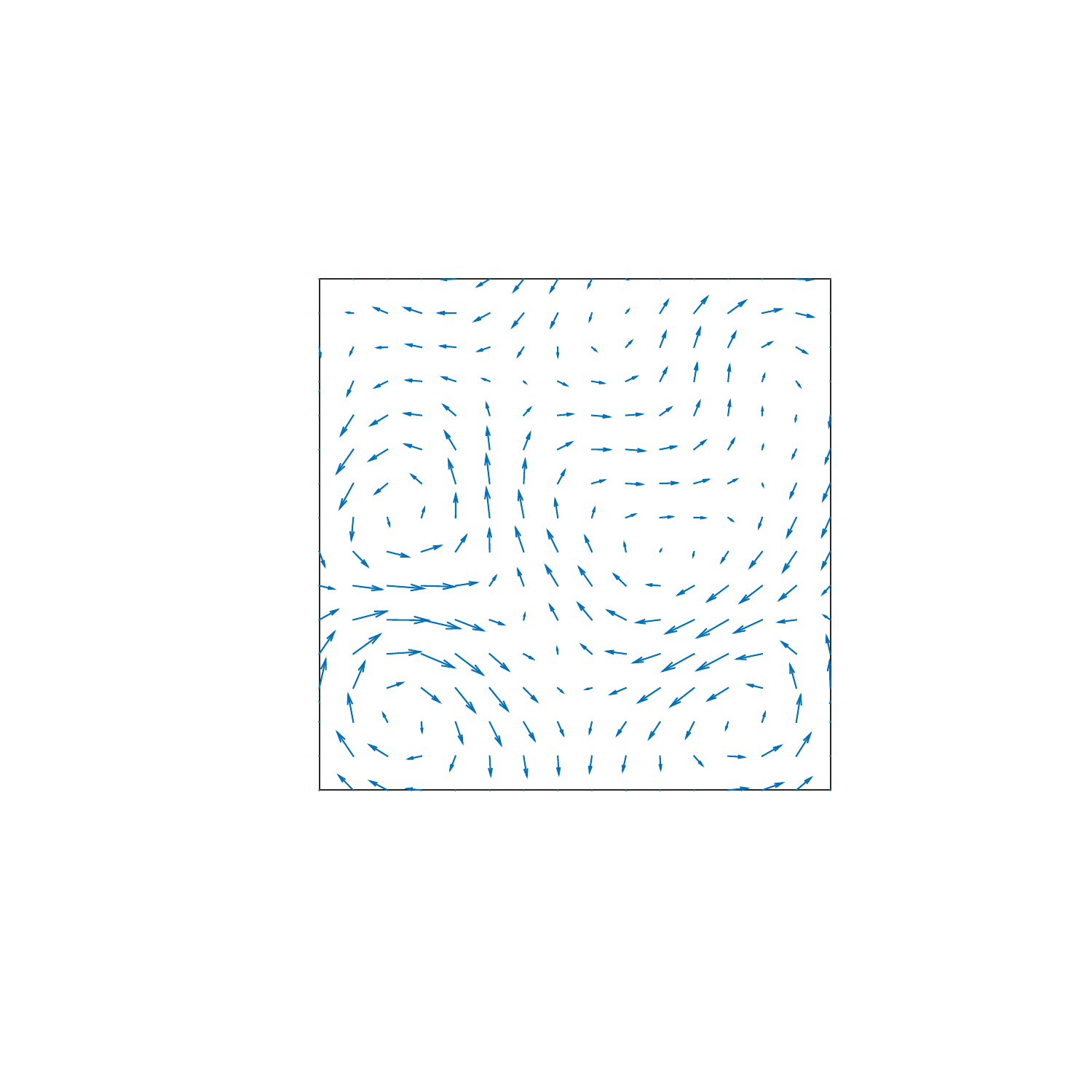}  &\hspace{-0.4cm}  \includegraphics[height=0.3\columnwidth]{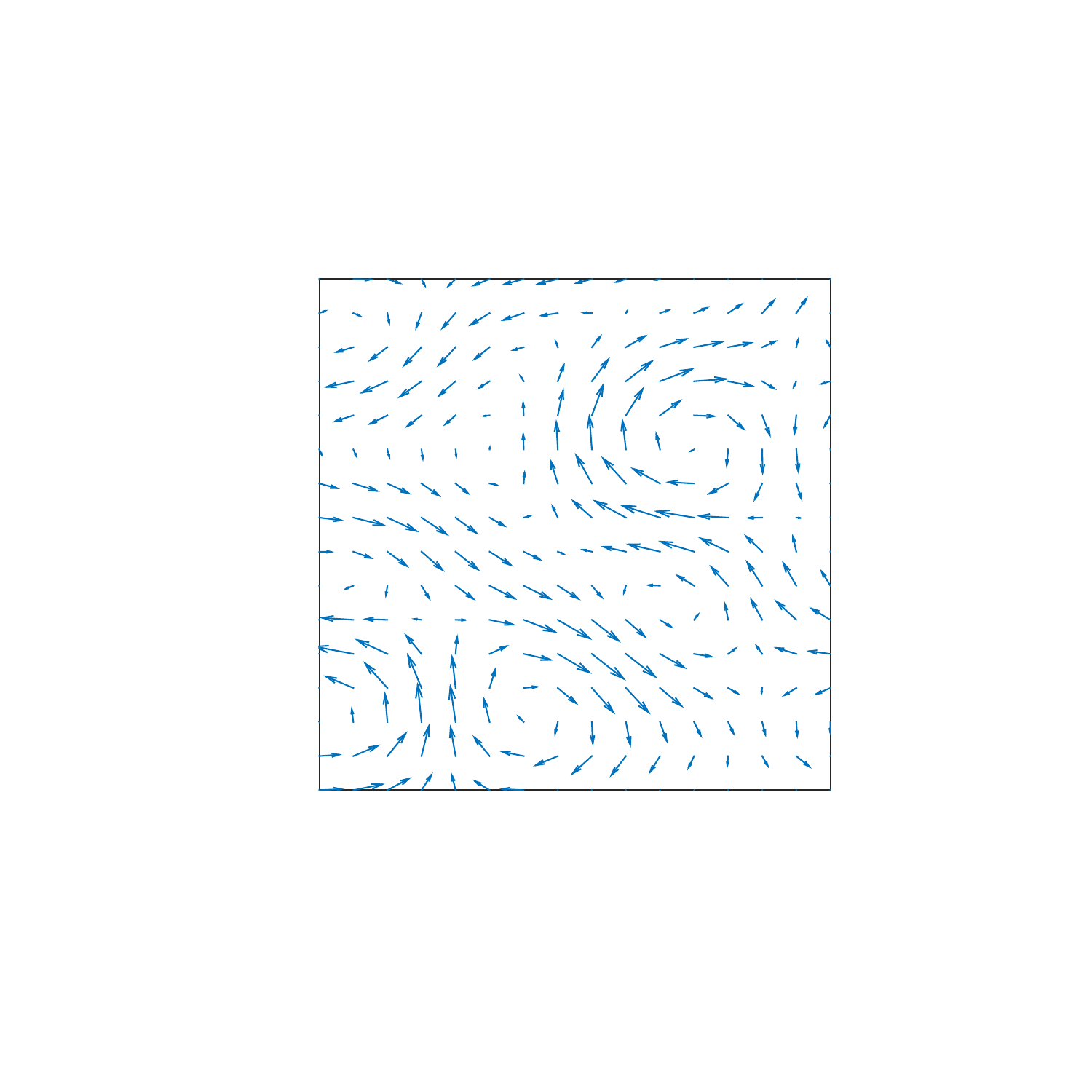} 
\end{tabular}
\caption{\small Three examples of snapshot $x_t(\thetaLearn_j)$:  maps of  absolute vorticity  and temperature   generated by Rayleigh-B\'enard convection (above); motion fields generated by the divergence-free fBm evolution model (below). \vspace{-0.cm}\label{fig:1}}\vspace{-0.cm}
\end{figure}

 The size of the training data is set to $m=90$ for the two high-dimensional systems. The experimental setup is as follows.
 \begin{itemize}
 \item For the Rayleigh-B\'enard convection model, 10 initial
conditions $\thetaLearn_j$ are sampled from a uniform distribution on an
hyper-cube  in $\Rr^5$ parametrizing  solutions of the Lorenz attractor~\cite{Lorenz63}.  Then using $\thetaLearn_j$ to initialize  the dynamic model, we compute trajectories for $t=1,\ldots,10$ (resulting in 100 states $x_t(\thetaLearn_j)$). 
Examples of $x_t(\thetaLearn_j)'s$ are displayed in Figure~\ref{fig:1}.
We design the test data in order to evaluate the capability of the reduced models to perform predictions for the Rayleigh-B\'enard convection model, which is known to be chaotic~\cite{{chandrasekhar2013hydrodynamic},Lorenz63}. Therefore, we set the  test data as the prolongation of the training data trajectories:  the 10 initial conditions  are  $\theta_j=x_{10} (\thetaLearn_j)$  and  trajectories
$x_t(\theta_j)$  for $t=1,\ldots,10$  are computed in the same way as for
the training data set.
%
%
% The test data  is  set as follows:   10 initial conditions  are set as  $\theta_j=x_{10} (\thetaLearn_j)$  and  trajectories
%$x_t(\theta_j)$  for $t=1,\ldots,10$  are computed in the same way as for
%the training data set. \remPH{challenging steup, comme ICASSP}
\item  For the divergence-free fBm evolution model, 100 noisy fBms vector fields $\thetaLearn_j$ are used as initial conditions. More precisely, fBms vector fields are drawn using the wavelet representation proposed in~\cite[Proposition 3.1]{heas2014self} with 18 fractional wavelet  coefficients and then corrupt  by a weak  additional  Gaussian white noise (peak signal-to-noise ratio of $130$). We then compute  trajectories for $t=1,2$ (resulting in 100 states $x_t(\thetaLearn_j)$)  using the quadratic dynamical model. Examples of $x_t(\thetaLearn_j)'s$ are displayed in Figure~\ref{fig:1}. The test data set is generated in the same way.\\
\end{itemize}

 \subsection{Results}  
\subsubsection{Analysis of  accuracy {\it vs.} rank of  approximation}\label{sec:QuantAnalysis}

  \begin{figure}[!t]
\centering
\vspace{-2.5cm}\begin{tabular}{cc}\vspace{-2cm}
\hspace{-1.5cm}  \includegraphics[height=0.7\columnwidth]{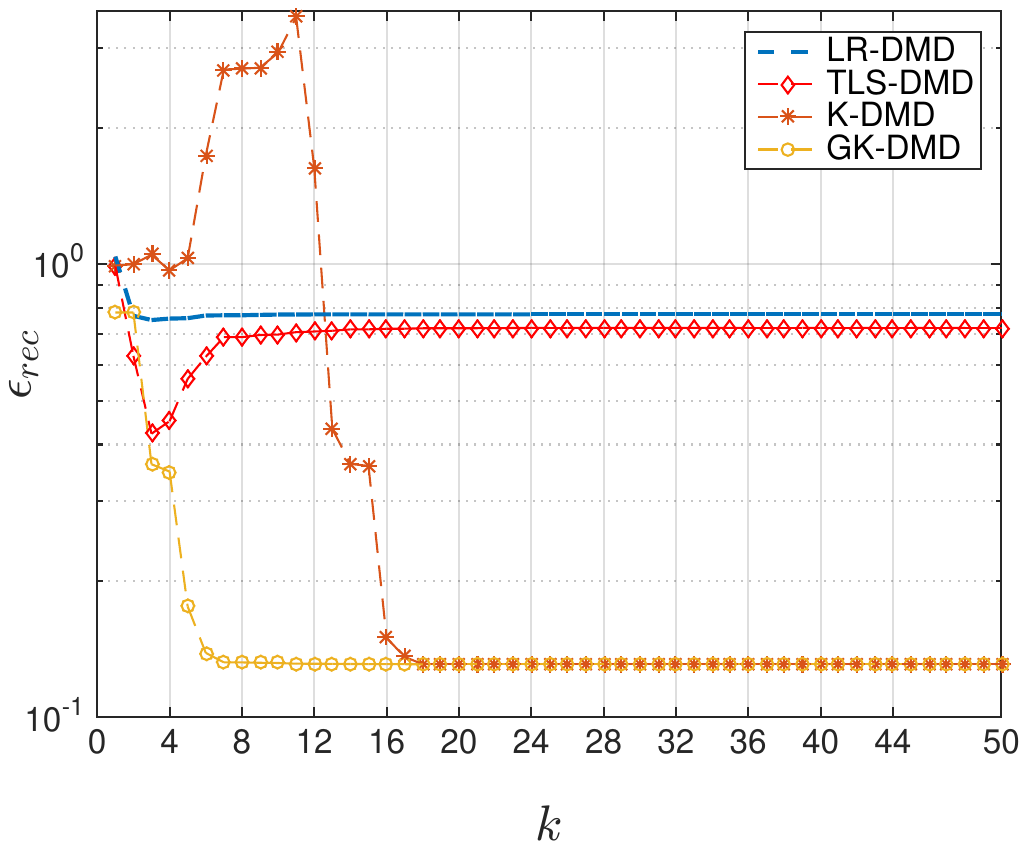}&\hspace{-1.5cm}  \includegraphics[height=0.7\columnwidth]{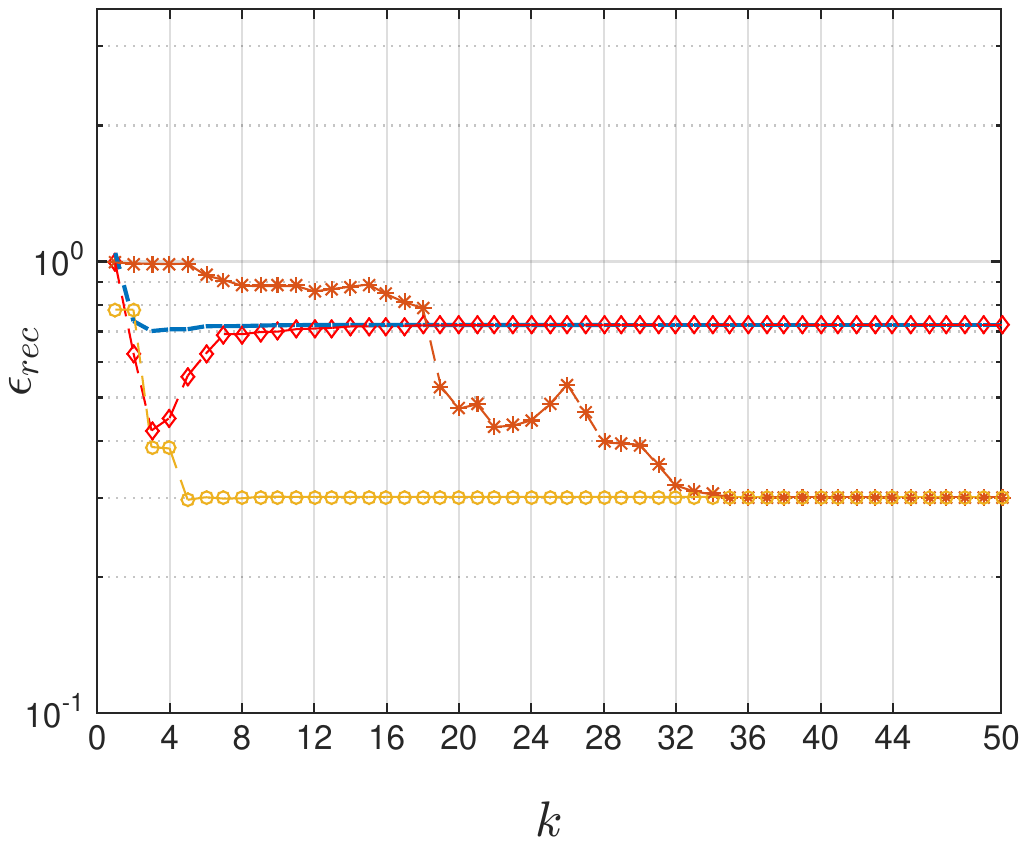}\vspace{-0.5cm} \\
%\hspace{-1.cm} \includegraphics[height=0.65\columnwidth]{./Figures/Eval_NoNoiseSequence5_half_log.pdf}  &\hspace{-0.4cm}  \includegraphics[height=0.65\columnwidth]{./Figures/EvalPredict_0NoiseSequence5_log.pdf} \vspace{-2cm}\\ 

\vspace{-0.25cm}
\end{tabular}
\caption{\small Evaluation of reconstruction error $\epsilon_{rec}$  as a function of   the  rank~$k$  for the approximation of a Rayleigh-B\'enard convective  system using Gaussian (left) and polynomial  (right) kernels. \vspace{-0.cm}\label{fig:new0}\label{fig:1_}}\vspace{-0.cm}
\end{figure}

 \begin{figure}[t!]
\centering
\vspace{-2.cm}\begin{tabular}{cc}
%\hspace{-1.cm} \includegraphics[height=0.45\columnwidth]{./Figures/logKernelTrain4.pdf}  &\hspace{-0.4cm}  \includegraphics[height=0.45\columnwidth]{./Figures/logKernelTest4.pdf} \\
%\hspace{-0.7cm}\includegraphics[height=0.45\columnwidth]{./Figures/gaussKernelTrain4.pdf}&
%\hspace{-0.25cm}\includegraphics[height=0.45\columnwidth]{./Figures/gaussKernelTest4.pdf}\vspace{-0.75cm}\\
\hspace{-0.25cm}\includegraphics[height=0.7\columnwidth]{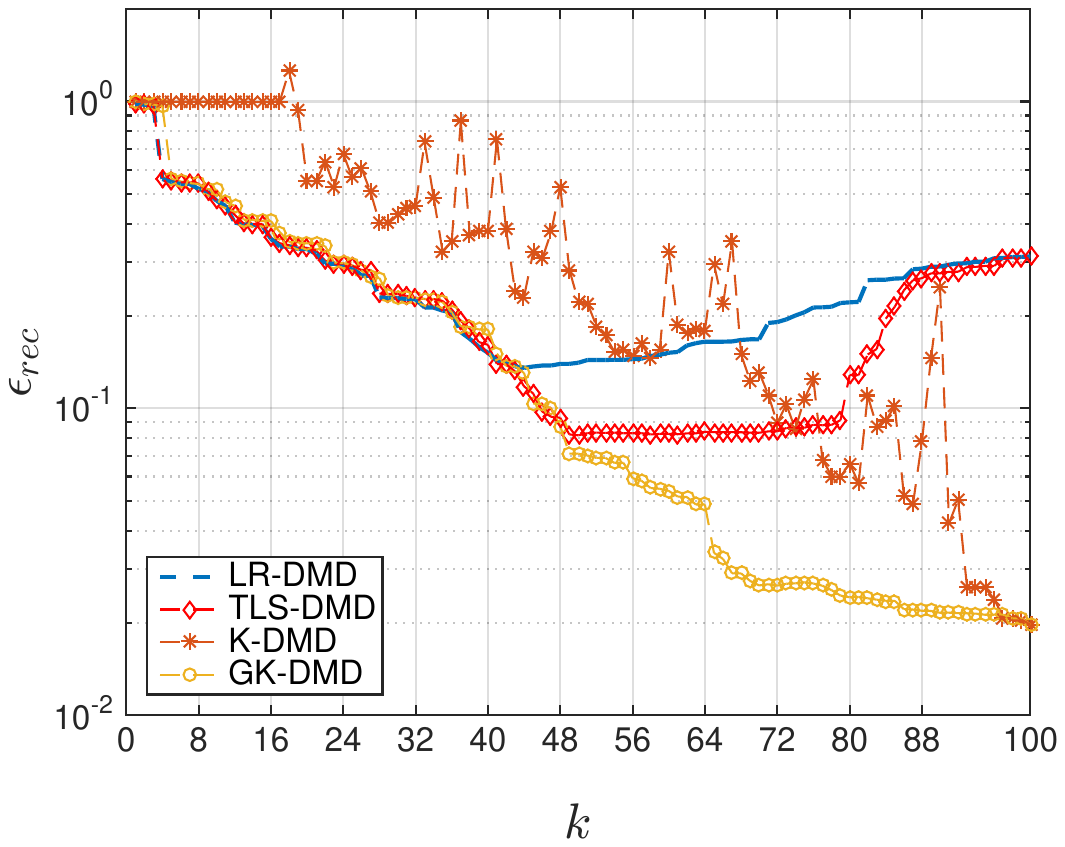}\vspace{-0.5cm}\\
%\hspace{-0.75cm} \includegraphics[height=0.45\columnwidth]{./Figures/poly2KernelTrain4.pdf}  &\hspace{-0.25cm}\includegraphics[height=0.45\columnwidth]{./Figures/poly2KernelTest4.pdf}\\
%\hline \\
%{Training}&{Testing}\\
\vspace{-2.25cm}
\end{tabular}
\caption{\small  Evaluation of the reconstruction error  $\epsilon_{rec}$ as a function of   the  dimension~$k$ for the approximation of a divergence-free fBm evolution model using 	a Gaussian kernel. \vspace{-0.5cm}\label{fig:0}}\vspace{-0.cm}
\end{figure}

The reconstruction error $\epsilon_{rec}$  is evaluated as a function of the reduced model dimension for the approximation of the two  high-dimensional systems. Plots comparing the four algorithms   are presented in Figure~\ref{fig:new0} and  Figure~\ref{fig:0}, respectively for the Rayleigh-B\'enard convection and for the divergence-free fBm evolution.  Overall, we observe that GK-DMD outperforms almost everywhere the other methods.  \vspace{-0.2cm}  \\

{\bf Rayleigh-B\'enard convection.}  We first discuss the results  shown in Figure~\ref{fig:new0} for the Gaussian kernel.    While K-DMD and GK-DMD perform similarly  for  $k\ge 18$,  GK-DMD exhibits for $k<18$ a clear gain in accuracy compared to the other methods  reaching almost an order of magnitude.  The  gain in accuracy  between  GK-DMD and   K-DMD may be due to the fact that  GK-DMD  computes exactly  the reduced model \eqref{eq:koopman1}, and in particular considers the low-rank solution  $A^\star_k$ instead of  the truncated least square solution $\hat A^{\ell s}_k$.
 Besides, as $\textrm{rank}(\AAA^*\AAA)=m$, \ie operator  $\AAA$ is full-rank,   similar performances of the two kernel-based methods  in the case where  $k\ge 18$  can be explained by the fact that the low-rank constraint becomes inactive
 (implying that $\hat A^{\ell s}_k=A^\star_k$),     $\Psi^{-1}$ is well approximated by a linear mapping and furthermore the $ \Psi^{-1} \zeta_j$'s are well represented in the span of $\YYY$. A lower value on the accuracy is reached around $k$ slightly greater than $5$, suggesting that the reduced model can explain up to $5$ components  in $\mathcal{H}$.  This lower bound on the error shows that the snapshots used for training have redundant information in $\mathcal{H}$ about the Rayleigh-B\'enard convection system and are insufficient to represent the wide variety of trajectories belonging to the solution manifold. Similar results are obtained with a polynomial kernel. Nevertheless, for  K-DMD and GK-DMD , the reconstruction error is higher with polynomials than for Gaussian kernels, revealing that their performance is kernel-dependent.  
The  problem of the kernel choice, and more generally the inference of a relevant mapping $\Psi$,  is an open question out of the scope of this paper. We refer the interested reader to recent works on this issue~\cite{Lusch2018DeepLF,yeung2019learning}.

Additionally, the reconstruction errors of GK-DMD,  LR-DMD and TLS-DMD are comparable for $k < 4$. However, for larger values of $k$, the performance of nonlinear methods is much better: the  error gain brought about by the use of kernels reaches around $93$\% for the Gaussian kernel and $73$\% for the polynomial kernel. More precisely, the accuracy of  LR-DMD and TLS-DMD reaches a lower bound around $k \simeq 4$ and then 
deteriorates as $k$ increases  or reaches an asymptote, suggesting data overfitting.  This undesirable effect will be discussed in the numerical experiences of  Section~\ref{sec:overfit}.\vspace{-0.3cm}\\

{\bf  Divergence-free fBm evolution}. We now turn to the reduction of our simplified turbulence model. Results  shown in Figure~\ref{fig:0} are in agreement with observations made in the case of the  Rayleigh-B\'enard convection. In particular the gain in accuracy in comparison to state-of-the-art methods reaches a peak of more than an order of magnitude at $k = 89$. However, we can point out two important differences with the previous results shown in Figure~\ref{fig:new0}. First,  while the GK-DMD error is a monotonic strictly  decreasing function with respect to the dimension $k$, K-DMD has a plateau up to   $k \sim20$ and then is chaotic   before reaching the GK-DMD performance near the point $k=m$.  The conclusion that can be drawn is that GK-DMD provides a preferable reduced model, in the sense that approximation becomes increasingly accurate  as $k$ grows. Second, LR-DMD and TLS-DMD operate in a very similar way as GK-DMD up to $k \sim 40$,   but the accuracy of these methods deteriorates considerably as $k$ increases.   As already observed for the Rayleigh-B\'enard model, overfitting at the learning stage  is likely to be the explanation to this  performance loss of these state-of-the-art algorithms. This issue is further discussed in Section~\ref{sec:overfit}. Finally, we can also note the continuous decrease of the error with $k$, even if it is slow, which suggests that the snapshots used for training better represent the solution manifold than for the Rayleigh-B\'enard system.
\vspace{-0.cm}%, and is in accordance with the numerical experiments we will discuss in Section~\ref{sec:overfit}.  

\begin{figure}[t!]
\centering
\vspace{-0.cm}
\begin{tabular}{c|cc}
{\footnotesize True absolute vorticity}& \includegraphics[height=0.08\columnwidth]{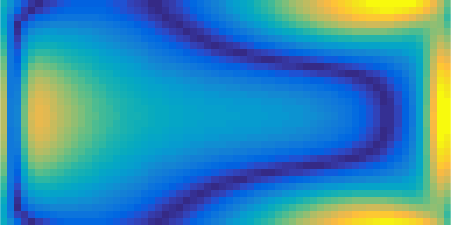} \\% &\includegraphics[height=0.1\columnwidth]{./Figures/XtrueAbs5_38_th_crop.png}   \\
&\includegraphics[width=0.175\columnwidth]{./Figures/colorbar_b}  % & \includegraphics[width=0.2\columnwidth]{./Figures/colorbar_th}   
\end{tabular}
\hspace{-0.2cm}\begin{tabular}{c | c c c c c}
{\footnotesize }&\multicolumn{1}{c}{\footnotesize{$k=5$}}&\multicolumn{1}{c}{\footnotesize{$k=10$}}&\multicolumn{1}{c}{\footnotesize{$k=15$}}&\multicolumn{1}{c}{\footnotesize{k=20}}\\
\hline \\ 
{\footnotesize LR-DMD}&\includegraphics[height=0.08\columnwidth]{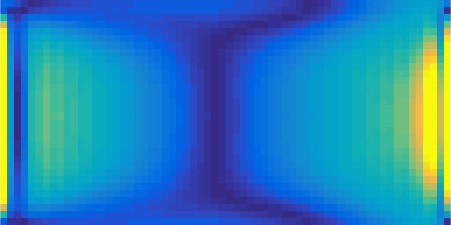}  &\includegraphics[height=0.08\columnwidth]{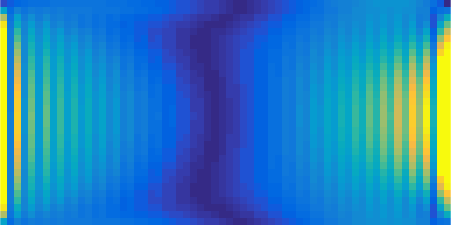}   &\includegraphics[height=0.08\columnwidth]{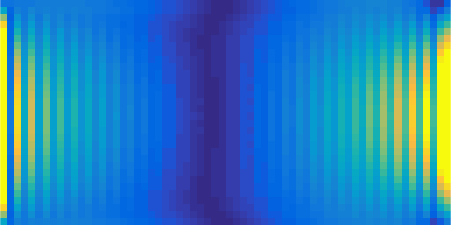} &\includegraphics[height=0.08\columnwidth]{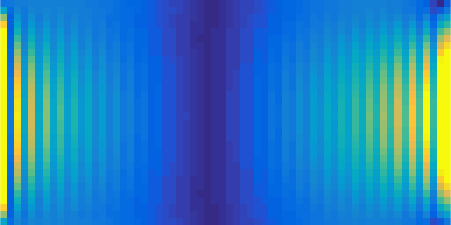}     \\
{\footnotesize TLS-DMD}&\includegraphics[height=0.08\columnwidth]{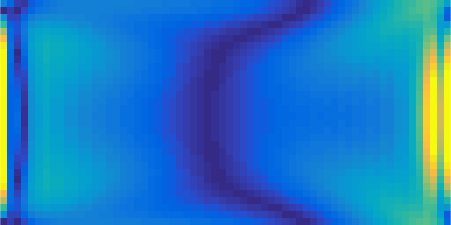}  &\includegraphics[height=0.08\columnwidth]{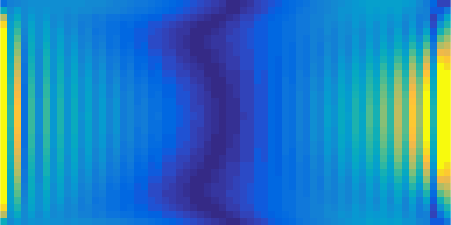}   &\includegraphics[height=0.08\columnwidth]{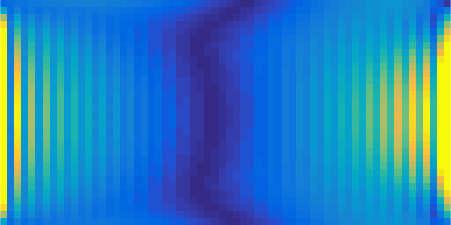} &\includegraphics[height=0.08\columnwidth]{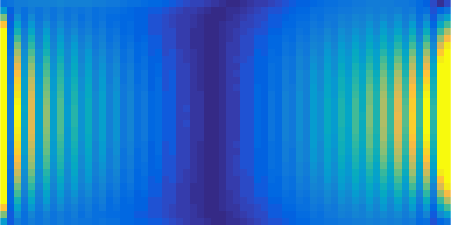}   \\
{\footnotesize K-DMD}&\includegraphics[height=0.08\columnwidth]{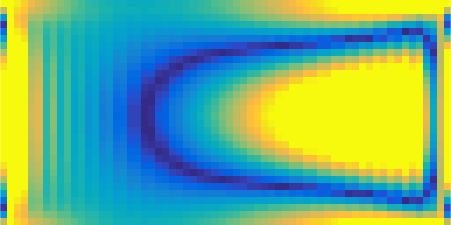}  &\includegraphics[height=0.08\columnwidth]{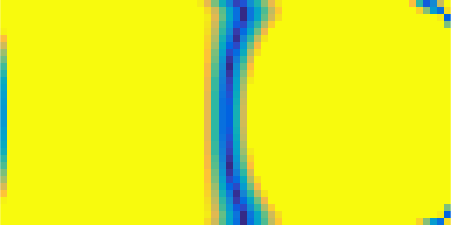}   &\includegraphics[height=0.08\columnwidth]{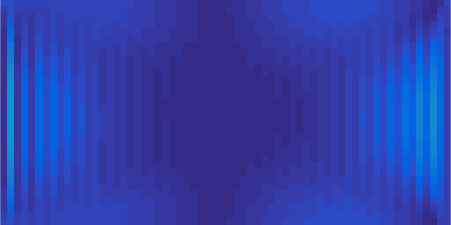} &\hspace{0.1cm}\includegraphics[height=0.08\columnwidth]{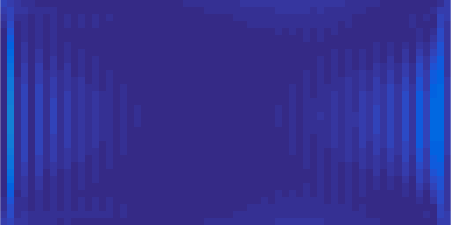}   \vspace{-1.25cm}\\
{\footnotesize GK-DMD}&\includegraphics[height=0.08\columnwidth]{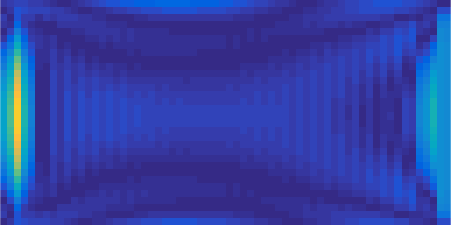}  &\includegraphics[height=0.08\columnwidth]{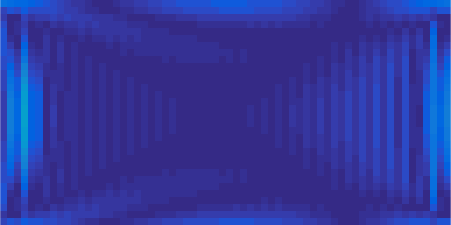}   &\includegraphics[height=0.08\columnwidth]{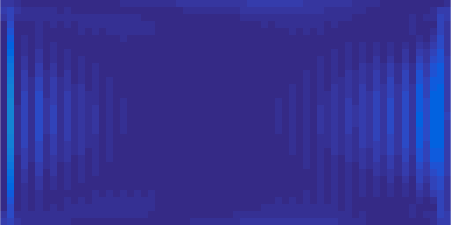} &\includegraphics[height=0.08\columnwidth]{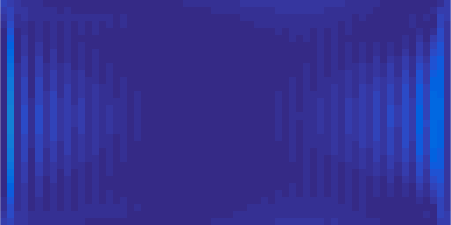}    &\includegraphics[width=0.0525\columnwidth]{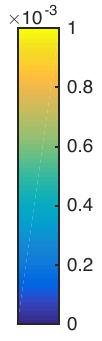}\vspace{-0.2cm}
\end{tabular}
\caption{\small Reconstruction error maps of the different reduced models for increasing value of $k$. The image above represents the absolute vorticity field $x_{2}(\theta)$, while each  image in the table below  represents the absolute vorticity  of the bi-variate  field ${ \tilde x_2( \theta)-x_{2}(\theta)}$, for a typical initial condition  $ \theta$. \vspace{-0.cm}\label{fig:new3}}\vspace{-0.cm}
\end{figure}

 \begin{figure}[t!]
\centering
\vspace{-0.cm}\begin{tabular}{cccc}
 \hspace{-1.25cm}$x_2(\theta)$ &\hspace{-1.25cm} LR-DMD error &\hspace{-1.25cm}TLS-DMD error&\\
\hline
\hspace{-1.25cm}\includegraphics[height=0.3\columnwidth]{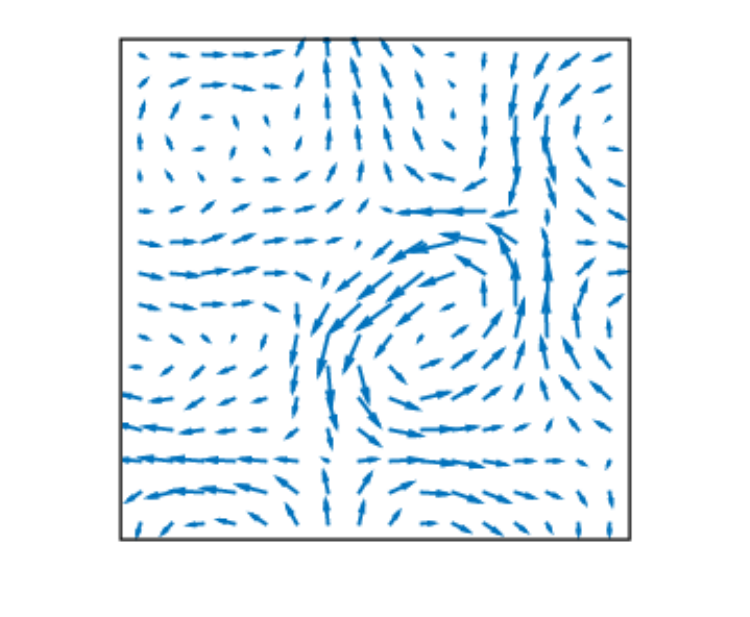}&\hspace{-1.25cm}\includegraphics[height=0.3\columnwidth]{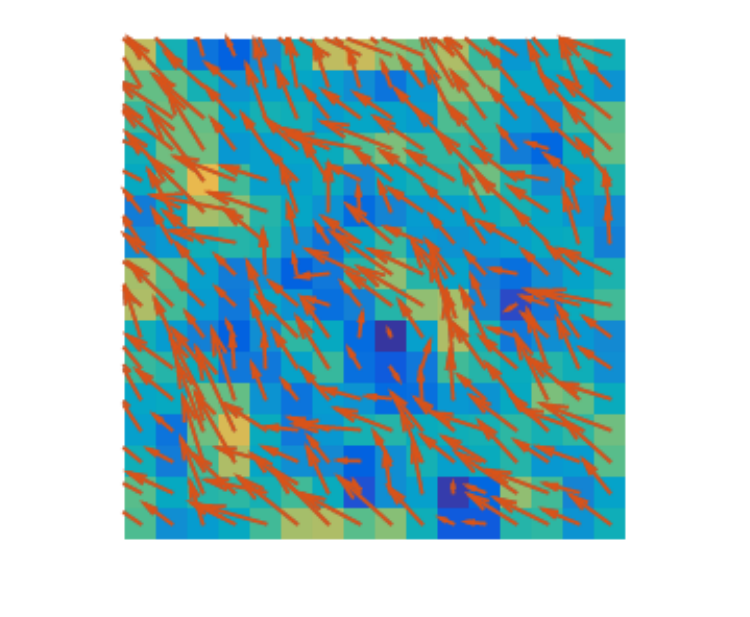}&\hspace{-1.25cm}\includegraphics[height=0.3\columnwidth]{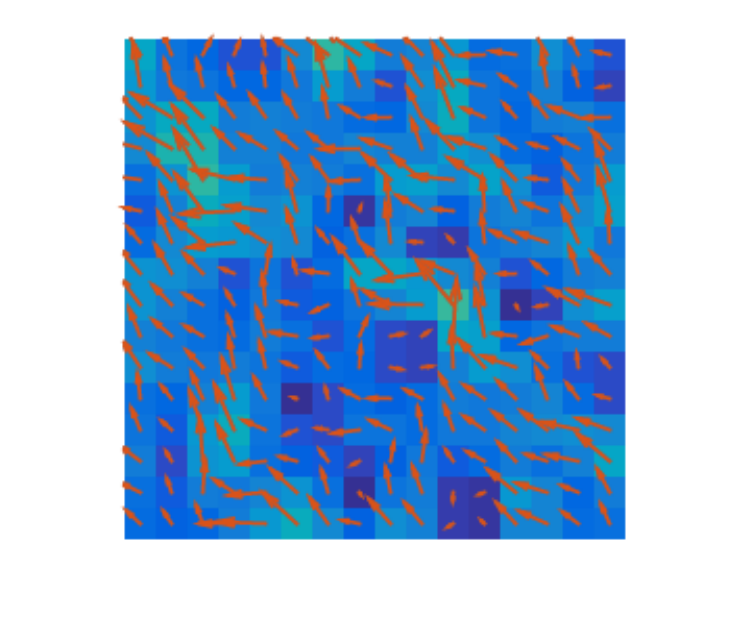}&\vspace{-0cm}\hspace{-0.75cm}\includegraphics[height=0.2\columnwidth]{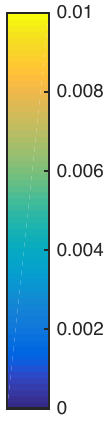} \vspace{-0.5cm}\\
\hspace{-1.25cm}&\hspace{-1.25cm}{K-DMD error}&\hspace{-1.25cm}{GK-DMD error }\\
\hline 
&\hspace{-1.25cm}   \includegraphics[height=0.3\columnwidth]{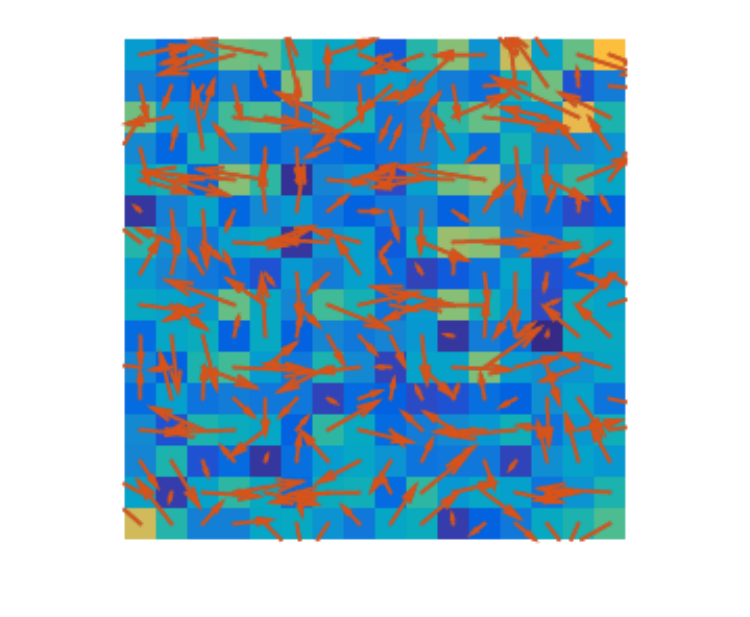} &\hspace{-1.25cm}\includegraphics[height=0.3\columnwidth]{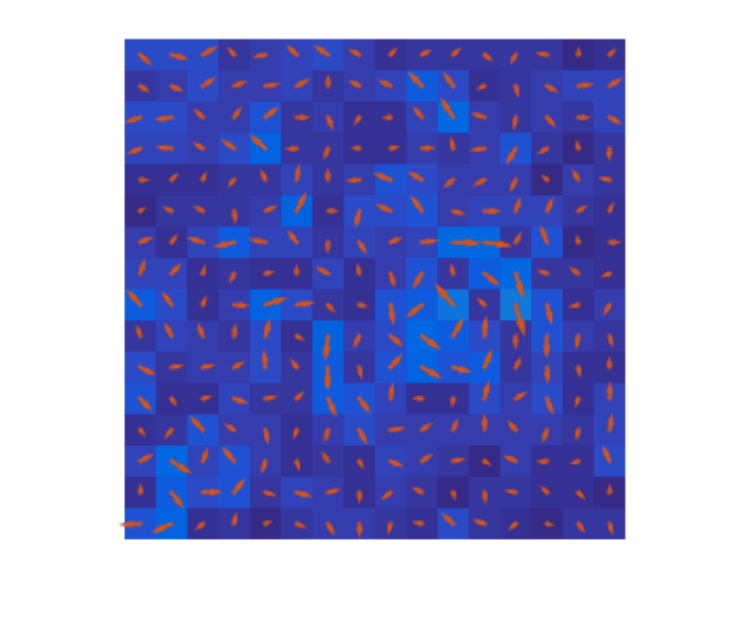}\\
\vspace{-1.cm}
\end{tabular}
\caption{\small  Reconstruction errors maps of the different ROM approximations  for $k=70$.  The vector field at upper left corner is related to the snapshot $x_2(\theta)$ for a typical initial condition  $ \theta$, while other vector fields enlarged at scale $20$:$1$ are related to the reconstruction error $\tilde x_2(\theta)- x_2(\theta)$ superimposed on the maps of  the absolute error  $|\tilde x_2(\theta)- x_2(\theta)|$. \vspace{-0.5cm}\label{fig:2}}
\end{figure}

\subsubsection{Analysis of error maps}\label{sec:QualAnalysis}
 To complement  the quantitative  evaluation performed in Section~\ref{sec:QuantAnalysis}, we  proceed to the visual inspection of the   spatial distribution of the error. Typical error maps  are shown in  Figure~\ref{fig:new3} and  Figure~\ref{fig:2}. Figure~\ref{fig:new3} displays the absolute vorticity  of the bi-variate error  field  ${ \tilde x_2( \theta)-x_{2}(\theta)}$ defined over the bi-dimensional grid, where $\tilde x_2(\theta)$ denotes the  approximation provided by  the algorithms for a given initial condition $\theta$. Error maps are displayed for increasing value of the dimension $k$.  The distribution of the error produced by K-DMD in Figure~\ref{fig:new3}  reveals that its chaotic behavior for $k<15$ is caused by the occurence of errors at large scales. Error maps obtained with the LR-DMD and TLS-DMD algorithms are very similar. Moreover they seem not to evolve significantly as $k$ increases, except for high frequency  appearing at $k \ge 10$.      The error maps for GK-DMD show that the decrease in error with respect to $k$ is related to  refinements occurring at increasingly finer scales. 
 Interestingly, the simplified turbulence model confirms this scale analysis of errors.  In Figure \ref{fig:2}, we observe that the large vortices of the flow are correctly approximated by GK-DMD for $k=70$. By contrast, GK-DMD does not reconstruct accurately   less energetic small vortices of the divergence-free fBm. Although TLS-DMD produces a reasonable error,  we remark that TLS-DMD and LR-DMD clearly fail at identifying some of the large structures of the flow. The inspection of error maps produced by K-DMD reveal large errors at all scales, and  its inaccuracy for reduced modeling of the  fractal structure of turbulence.

\subsubsection{Robustness to overfitting}\label{sec:overfit}

  \begin{figure}[t!]
\centering
\vspace{-0.95cm}\begin{tabular}{c}\vspace{-2cm}
%\hspace{-1.cm} 
 \includegraphics[height=0.7\columnwidth]{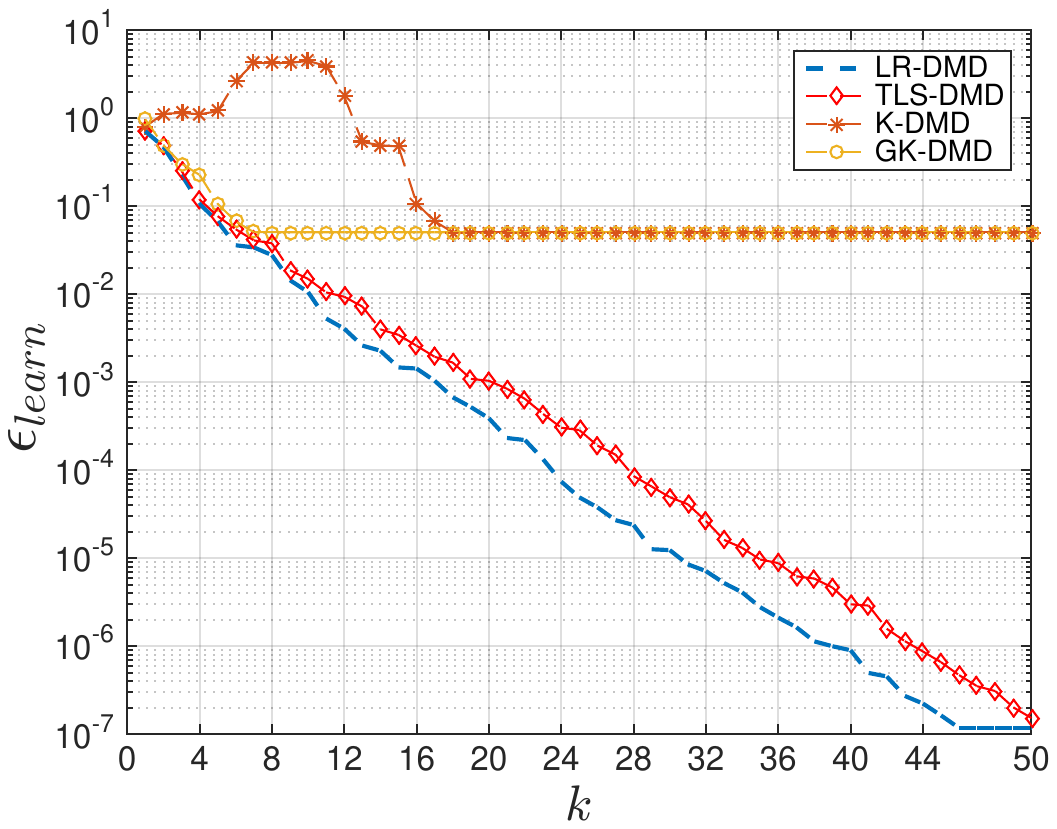}%&\hspace{-0.8cm}  \includegraphics[height=0.825\columnwidth]{./Figures/Evaltrain_0NoiseSequence5_poly.pdf}\vspace{-0.5cm} \\
%\hspace{-1.cm} \includegraphics[height=0.65\columnwidth]{./Figures/Eval_NoNoiseSequence5_half_log.pdf}  &\hspace{-0.4cm}  \includegraphics[height=0.65\columnwidth]{./Figures/EvalPredict_0NoiseSequence5_log.pdf} \vspace{-2cm}\\ 

\vspace{-0.6cm}
\end{tabular}
\caption{\small Evaluation of learning error $\epsilon_{\text{learn}}$  as a function of   the  dimension~$k$ of the reduced-model approximation, for a Gaussian kernel. \vspace{-0.cm}\label{fig:new02}}\vspace{-0.cm}
\end{figure}

We pointed out in the previous analysis that LR-DMD and TLS-DMD might be prone to overfitting, while kernel methods seem to be more robust. We now discuss this point by evaluating the learning error $\epsilon_{\text{learn}}$  as a function of the reduced model dimension $k$ for the approximation of  Rayleigh-B\'enard convection. Curves illustrating the learning error attained by the four algorithms   are presented in Figure~\ref{fig:new02}.

 The  plot of Figure~\ref{fig:new02} shows that  LR-DMD, TLS-DMD or GK-DMD   perform  good and similarly in terms of learning error up to  $k\sim 7$ (while K-DMD fails completely). From then, the learning error is continuously reduced as $k$ grows  by  LR-DMD and TLS-DMD, while GK-DMD reaches an asymptote (the same as K-DMD from $k=17$).  We point out that the gain of LR-DMD and TLS-DMD compared to GK-DMD  reaches  more than five orders of magnitude at $k=50$. Noticing that, on the contrary to this continuous decrease of the learning error, the reconstruction error of  LR-DMD and TLS-DMD in Figure~\ref{fig:new0} stagnates for $k\ge 7$, we conclude that these methods overfit the training samples. The observed robustness of GK-DMD and K-DMD  in our experiments  is in accordance with empirical results obtained for K-DMD in~\cite{williams2014kernel}: in their numerical simulations the authors showed that immersing the dynamics in the high-dimensional space  $\mathcal{H}$ enhanced the  eigen-values estimates by lowering their variance and discarding neglectible ones.  \vspace{-0.1cm} %Similar conclusions can be drawn for the case of a polynomial kernel  displayed in the right plot of Figure~\ref{fig:new02}.

\subsubsection{Robustness to noise}

  \begin{figure}[t!]
\centering
\vspace{-0.9cm}\begin{tabular}{cc}
\hspace{-1.cm} \includegraphics[height=0.7\columnwidth]{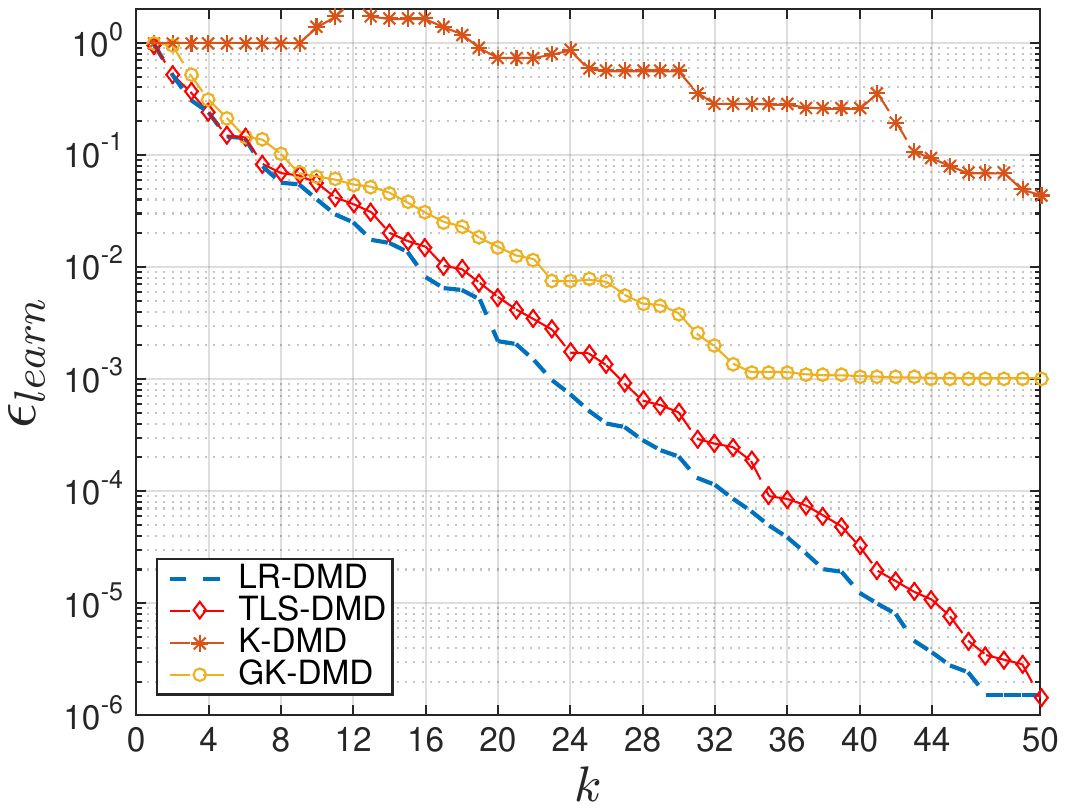}  &\hspace{-1.5cm}  \includegraphics[height=0.7\columnwidth]{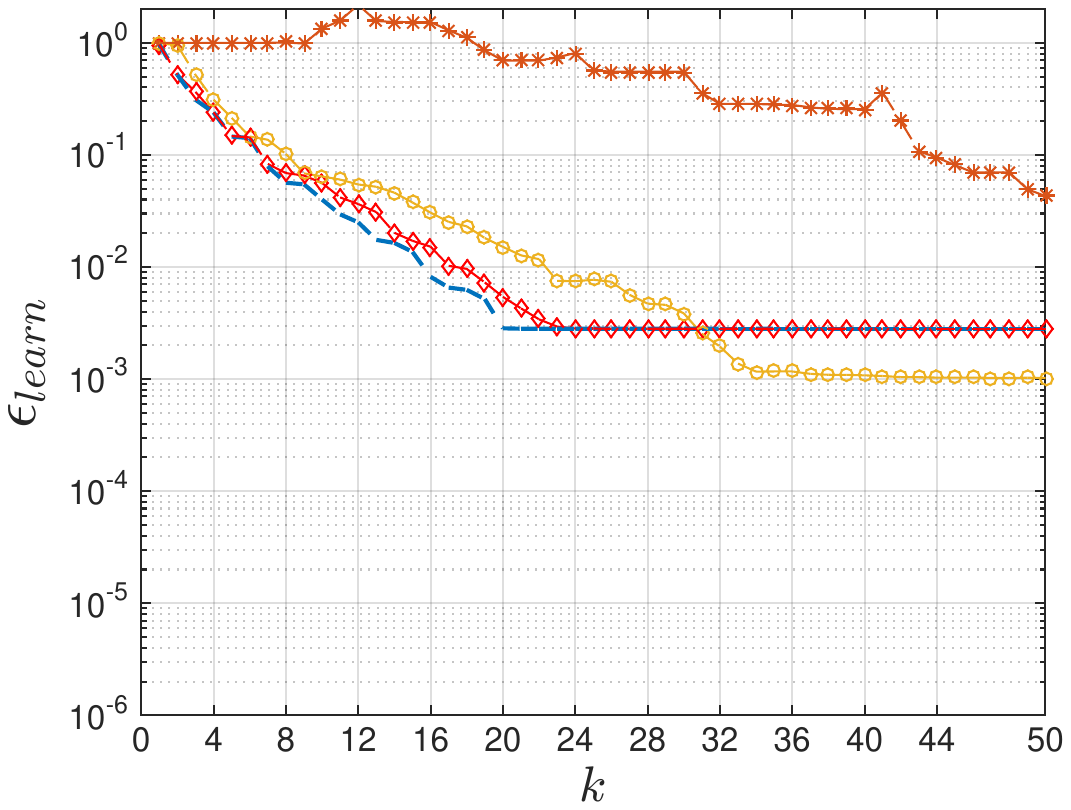} \vspace{-4.5cm}\\  
\hspace{-1.cm} \includegraphics[height=0.7\columnwidth]{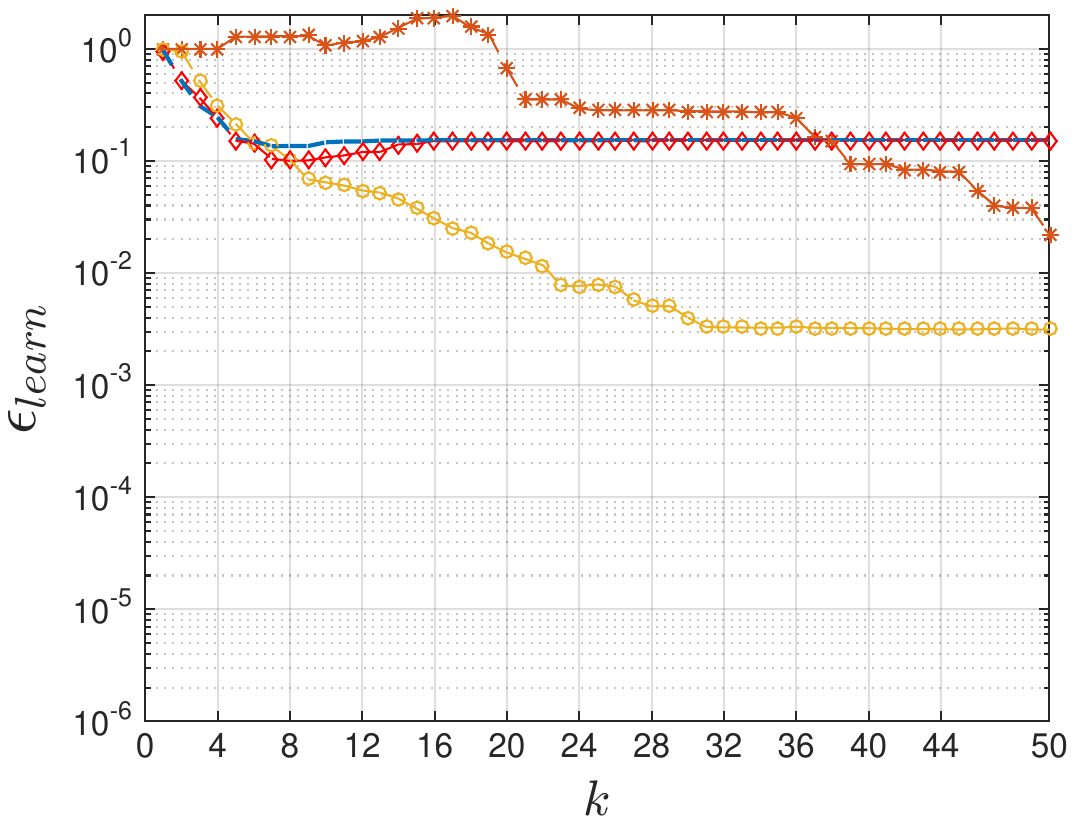} &\hspace{-1.5cm}  \includegraphics[height=0.7\columnwidth]{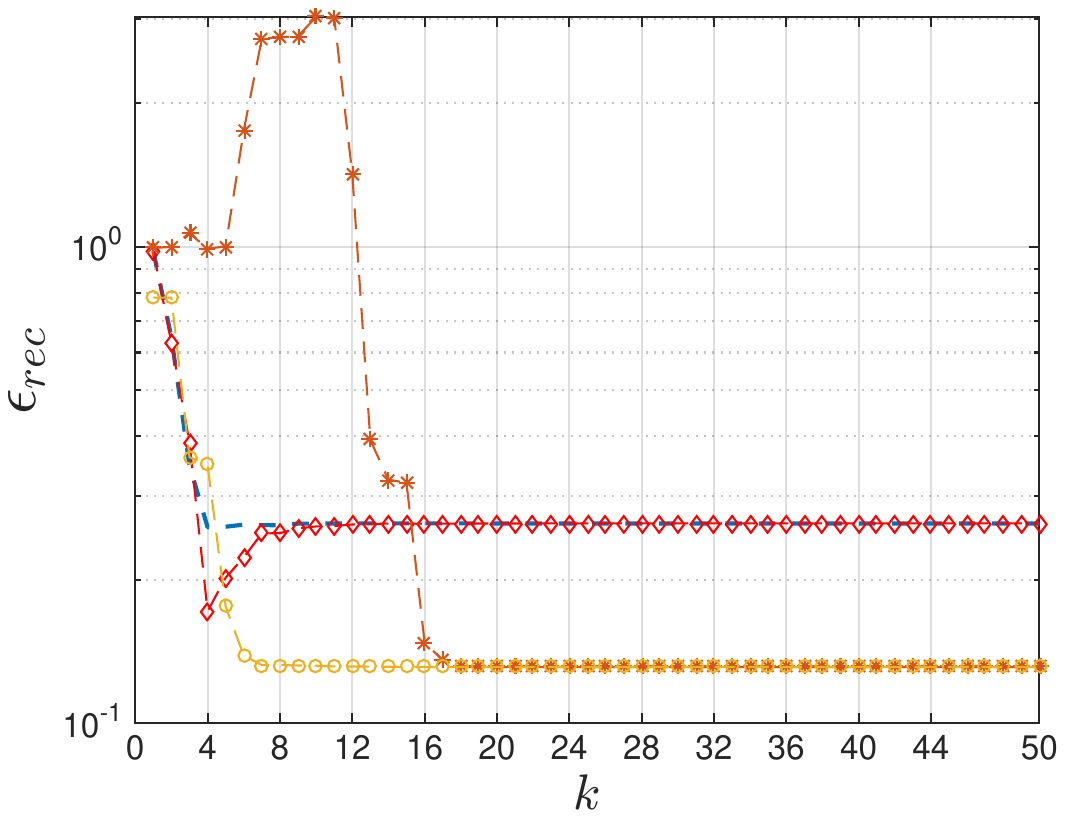} 
\vspace{-2.5cm}
\end{tabular}
\caption{\small  Evaluation of  errors $\epsilon_{\text{learn}}$  and $\epsilon_{rec}$ as a function of   the  dimension~$k$ of the reduced-model approximation using a Gaussian kernel and  different levels of additive Gaussian noise: $\epsilon_{\text{learn}}$ for  signal-to-noise ratio equal to $\infty$  (upper left), 54  (upper right) or 34 (lower left); $\epsilon_{rec}$ for  signal-to-noise ratio equal to 54 (lower right). \vspace{-0.cm}\label{fig:new1}}\vspace{-0.5cm}
\end{figure}

We assess the robustness  of the reduced models  by evaluating the learning error $\epsilon_{\text{learn}}$ in the noiseless case or in the case of an additive Gaussian noise of increasing variance.  We consider  the approximation of  Rayleigh-B\'enard convection. Figure~\ref{fig:new1} presents the learning error as a function of the reduced model dimension. We observe that performances of kernel-based methods, and particularly  GK-DMD, seem to be weakly  sensitive to the power of the additive noise.  On the contrary, we remark a significant increase of the learning error for LR-DMD or TLS-DMD in the presence of noise. This deterioration is accentuated by the noise power.  However,  comparing the curves in Figure~\ref{fig:new0} and Figure~\ref{fig:new1}, we notice that the reconstruction error  $\epsilon_{rec}$ of LR-DMD and TLS-DMD is lower in the presence of noise, and thus that an additive noise in the learning phase increases the prediction capabilities of these two reduced models. This behavior is likely to be the consequence of overfitting.  On the contrary,  in the case of a noisy setting, the analysis of the reconstruction error  confirms that the reduced model learned by GK-DMD is relevant for prediction. In particular,  results in Figure~\ref{fig:new1}  are consistent with those  obtained in  a noiseless setting   in Figure~\ref{fig:new0}. In particular, GK-DMD outperforms  in terms of accuracy the other methods for $k \le 16$.

  \section{Conclusions}\label{sec:conclusion}\vspace{-0.cm}
This work presents a data-driven algorithm for the tractable approximation of a linear low-rank operator characterizing dynamics embedded in a RKHS. Taking advantage of  the reproducing kernel property, the algorithm computes off-line a low-dimensional spectral representation of the  solution of a low-rank constrained optimization problem, and then solves on-line a minimum distance problem which yields the sought approximation.  By contrast to existing algorithms, the proposed reduced model  exhibits a low computational complexity and requires mild assumptions.  
Numerical simulations illustrate  the gain in accuracy allowed by the  proposed algorithm in the case of a meteorological system and a  turbulence model. \vspace{-0.cm}

%An open question raised by this work is the choice of the kernel, which is problem dependent. An interesting perspective is the use of data-driven learning strategies.    

%The idea underlying the method is to rely on a low-dimensional representation of the left and right eigen-vectors of the solution $A^\star_k$ of this problem. The representation enables to obtain: {\it i)} closed-form  eigen-functions of the Koopman operator approximation, expressed in terms of the left eigen-vectors of  $A^\star_k$; {\it ii)} the reduced model approximation $\tilde x_{\TT}(\theta)$ for a given initial condition $x_1=\theta$, given in terms of the right eigen-vectors of  $A^\star_k$ and exploiting the particular structure of the different kernels. 
%
%
%Based on this material, the proposed algorithm called GK-DMD is presented   and  particularisations of the method are presented respectively for polynomial, Gaussian and logarithmic kernels. 
% The algorithm presents a complexity in $\mathcal{O}(m(p+m^2))$ which is adequate for the case $p\ll n$, because of the  linearity with respect to the ambient dimension $p$ and independence of the functional space dimension $n$. 
%
% In particular, Finally, we also evaluate the influence of several features on the algorithm performances. In particular we highlight the importance of the kernel, which should be   carefully chosen according to the data. 

\appendix
\section{The K-DMD algorithm}\label{sec:kdmd}

Interested readers may find in this section a precise description of  the K-DMD algorithm introduced in~\cite{williams2014kernel}. K-DMD is exposed in Algorithm~\ref{algo:00}. 
We discuss hereafter its different steps and then make some comments on its complexity.

 First, K-DMD  computes matrices in step 1), relying on the kernel-trick. Next, in step 2), 3) and 6), K-DMD relies  on the assumption that $ A^{\star}_k=\hat A^{\ell s}_k$ and that we can identify  the left eigen-vectors of   $\hat A^{\ell s}_k$  as $\{\xi_{i}=U_{\AAA}   \tilde \xi_{i} \}_{i=1}^m$ and its eigen-values as $\{ \lambda_{i}=\tilde \lambda_{i}\}_{i=1}^m$. Here,  $\{\tilde \xi_i\}_{i=1}^m$ and  $\{\tilde \lambda_{i}\}_{i=1}^m$ denote the   right eigen-vectors  and eigen-values of the matrix  $R\, (\BBB^*  \AAA)\, R^*\in \Rr^{m \times m}$ with $R= \Sigma_{\AAA}^\dagger V_{\AAA}^*.$ % computed by the K-DMD algorithm. 
As shown by simple algebraic manipulations and using  Proposition \ref{prop:3.3}, the previous identification of eigen-vectors and eigen-values holds in the particular case where   $k=m$ (assumption {\it i}) and $\AAA$ is full rank (assumption {\it ii}). 
%we have the equality $(\tilde A_{\ell,m})^*=R\, (\BBB^*  \AAA)\, R^*$. It is straightforward to check that the equality holds if  
In this particular case, using the fact that $ U_{\AAA} =  \AAA  R^* $, 
 the $i$th   eigen-function approximation $ \varphi_i(\theta)$ can be evaluated at any point  $\theta \in \Rr^p$ as 
$
 \varphi_i(\theta)=  \langle \xi_i, \Psi(\theta) \rangle_{\mathcal{H}}= %(\Psi(\theta)^*  \AAA) R^* \tilde \xi_i
 \tilde \xi_i^* R \AAA^*  \Psi(\theta).\vspace{-0.cm}
$

\vspace{-0.cm}\begin{algorithm}[!h]
\begin{algorithmic}[0]
\State $\bullet$ \textbf{Off-line}

\begin{algorithmic}[0]
\State {\bf Inputs:}  the $x_t(\thetaLearn_i)$'s 
%\State 1)  Form matrix $\AAA$ and $\BBB$ as defined in \eqref{eq:matrixAB}. %, we obtain are able to compute $\mathbf{w}\in \Rr^{m \times m}$ satisfying 
\State 1) Compute matrices   $\AAA^*  \AAA$, $\BBB^*  \AAA  $ in $\Rr^{m \times m}$ with the kernel trick.%=U_{\hat\Q}\Sigma_{\hat\Q} V_{\hat\Q}^* $.%\ie and  the SVD of matrix $\AAA$ %, we obtain are able to compute $\mathbf{w}\in \Rr^{m \times m}$ satisfying 
%\begin{align*}
%\AAA&= U_a \,\begin{pmatrix}\textrm{diag}({\sigma_a}^{\frac{1}{2}})\\ 0_{n-\ell}\end{pmatrix}    V_a ^* .
%\end{align*}
%\State 2)  Define matrix   $D_k=U_{\hat\Q}^* \hat \R_k V_{\hat\Q}\Sigma_{\hat\Q} \in \Rr^{m\times m}$.
\State 2) Get $(V_{\AAA} , \Sigma_{\AAA})$ by eigen-decomposition of  $\AAA^*   \AAA$. 
%\State 4) Compute the projector $ \mathbb{P}_{\AAA^*}= %\AAA^\dagger\AAA=
%V_{\AAA}\Sigma_{\AAA}\Sigma_{\AAA}^\dagger V_{\AAA}^*$.  
%\State 3) The $i$th DMD mode corresponding to the eigen-value $\lambda_i$ is then given by 
%$$ \phi_i=.$$
%\State 3) Compute $\BBB^*\BBB$ where $\ZZZ$ is given by \eqref{eq:defZZZ}, given matrix $\BBB^* \BBB$.% and projector $ \mathbb{P}_{\AAA^*}= V_{\AAA}\Sigma_{\AAA}\Sigma_{\AAA}^\dagger V_{\AAA}^*$.
%\State 3) Get $(V_{\BBB} , \Sigma_{\BBB})$ by  eigen-decomposition of  $\BBB^*  \BBB$ . 
\State 3)  Compute  the eigen-vectors $\{ \tilde \xi_i  \}_{i=1}^m$ and eigen-values $\{\tilde \lambda_i \}_{i=1}^m$ of $R\, \BBB^*  \AAA\, R^*$, with $R= \Sigma_{\AAA}^\dagger V_{\AAA}^*$.
\State 4) Compute the pseudo-inverse $((\tilde \xi_1\cdots \tilde \xi_m)^*)^\dagger$.
\State \textbf{Outputs}: $R$, $ \tilde \xi_i$'s, $\tilde \lambda_i$'s and  $((\tilde \xi_1\cdots \tilde \xi_m)^*)^\dagger$
\end{algorithmic}
\end{algorithmic}
\begin{algorithmic}[0]
\State $\bullet$ \textbf{On-line}
\begin{algorithmic}[0]
\State \textbf{Inputs}:  off-line outputs and a given $\theta$
\State 5) Compute  $ \AAA^*  \Psi(\theta)$ in $\Rr^{m}$ with the kernel trick.
\State 6) Estimate  eigen-functions $\{  \varphi_i(\theta)\}_{i=1}^m$ using \eqref{eq:eigfuncApprox0}. %=\tilde \xi_i^* R \AAA^* \Psi(\theta).$$ 
\State 7) Estimate    eigen-modes $\{\mu_i\}_{i=1}^m$ using {\eqref{eq:eigmodeApprox}}. 
\State 8) Compute  $\tilde x_{\TT}(\theta)$ using \eqref{eq:eigenmoderepre} with  $\nu_{i,{\TT}}=\tilde \lambda_i^{{\TT}-1}  \varphi_i(\theta)$. 
\State \textbf{Output}: $\tilde x_{\TT}(\theta)$.
\end{algorithmic}
\end{algorithmic}
\caption{: K-DMD   \cite{williams2014kernel}\label{algo:00}} 
\end{algorithm}\vspace{0.1cm}

Then, the algorithm approximates  the   $\mu_j$'s in  reduced model \eqref{eq:eigenmoderepre}, with the underlying assumption that $\Psi^{-1}$ is linear (assumption {\it iii}). 
In steps 7), K-DMD identifies the  $\mu_j$'s  as the minimizers in $\Cr^p$ of the square of the $\ell_2$-norm of the reduced model error  
 %\begin{equation}\label{eq:recl2error}
 $\|x_{t+1}(\thetaLearn_i)-\sum_{j=1}^m   \mu_j \lambda_j \langle \xi_j , \Psi(x_t(\thetaLearn_i)) \rangle_\mathcal{H} \|_2^2$
 %,\vspace{-0.2cm}
 %\end{equation}
 for any data pair $(x_t(\thetaLearn_i),x_{t+1}(\thetaLearn_i))$ satisfying \eqref{eq:model_init}.  The authors thus implicitly assume that the eigen-modes belong to a sub-space in the span of $\YYY$  (assumption {\it iv}). 
 We notice  that by using \eqref{eq:eigfuncApprox0}, where $\Psi(\theta)$ is substituted by  $\Psi(x_{t}(\thetaLearn_i))$ for $i \in 1,\ldots, \NNPrim$ and $j \in 1,\ldots ,\TTPrim-1$, we have for $i=1,\ldots, m$ that
$$
%\xi_i^* \AAA= \tilde \xi_i^*  \Sigma_{\AAA} V_{\AAA}^*,\\
\begin{pmatrix} \langle \xi_i, \Psi(x_{1}(\thetaLearn_1))\rangle_{\mathcal{H}} \cdots  \langle \xi_i, \Psi(x_{\TTPrim-1}(\thetaLearn_\NNPrim))\rangle_{\mathcal{H}} \end{pmatrix}= \tilde \xi_i^*  \Sigma_{\AAA} V_{\AAA}^*.
$$
The estimated $\hat \mu_j$'s   are thus  rewritten as the  solution of \vspace{-0.1cm}
  \begin{align*}\argmin_{\mu_1,\ldots,\,\mu_m }\| \YYY- \begin{pmatrix}\mu_1& \cdots &\mu_m \end{pmatrix} \textrm{diag}(\tilde \lambda_{1}\cdots \tilde \lambda_{m} )(\tilde \xi_1\cdots  \tilde \xi_m)^* \Sigma_{\AAA} V_{\AAA}^* \|_{\mathcal{HS}},
 \end{align*}
where we remark that the Hilbert-Schmidt norm $\| \cdot \|_{\mathcal{HS}}$ boils down here to the Frobenius norm of a matrix in $ \Rr^{p \times m}$.  %Rewritten in matricial form using \eqref{eq:intermediaire0}, 
 A minimizer of this least-square optimization problem  is \vspace{-0.4cm}
% \begin{align}\label{eq:optim00}\argmin_{\mu_1,\ldots,\mu_m }\| \XXX- \begin{pmatrix}\mu_1& \cdots &\mu_m \end{pmatrix}\Xi^{-1}_m \AAA\|_F^2,
% \end{align}
% or 
% %{We remark that, assuming $\tilde \Xi_m^{-1}$ is full rank, we are searching for the minimal energy solution}. 
%
%%\begin{align}\label{eq:eigmodeApprox}
%% \begin{pmatrix}\mu_1& \cdots &\mu_m \end{pmatrix} \approx \XXX
%%% \begin{pmatrix}\xi_1^* \Psi(x_1^1)& \cdots &  \xi_1^* \Psi(x_1^N) \\ \vdots & & \vdots \\\xi_k^* \Psi(x_1^1)& \cdots &\xi_k^* \Psi(x_1^N)   \end{pmatrix} 
%%R^* ((\tilde \Xi_m)^{-1})^\dagger,
%%\end{align}
%%or 
\begin{equation}\label{eq:eigmodeApprox}
 \begin{pmatrix}\hat \mu_1& \cdots &\hat \mu_m \end{pmatrix} = \YYY
% \begin{pmatrix}\xi_1^* \Psi(x_1^1)& \cdots &  \xi_1^* \Psi(x_1^N) \\ \vdots & & \vdots \\\xi_k^* \Psi(x_1^1)& \cdots &\xi_k^* \Psi(x_1^N)   \end{pmatrix} 
R^* ((\tilde \xi_1\cdots \tilde \xi_m)^*)^\dagger  \textrm{diag}(\tilde \lambda_{1}^\dagger\cdots \tilde \lambda_{m}^\dagger ).
\end{equation}
 % \ie elements in the set $\{x_t(\theta_i)\}_{t=2,i=1}^{T,N}$ form the columns of matrix $\YYY$. 
Finally, using the identification $\{ \lambda_{i}=\tilde \lambda_{i}\}_{i=1}^k$, the $k$ first eigen-modes estimates $\{\hat \mu_i\}_{i=1}^k$ in \eqref{eq:eigmodeApprox}  with the $\varphi_i(\theta)$'s given by \eqref{eq:eigfuncApprox0} fully parametrize \eqref{eq:eigenmoderepre}, yielding the approximation 
$
 \tilde x_{\TT}(\theta)= \sum_{i=1}^{k}   \lambda_i^{t-1}  \varphi_i(\theta)\hat \mu_i.\vspace{-0.3cm} \\
$

 % with $( \Xi^{-1}_m)^*=( \xi_1\cdots  \xi_m)\in \Cr^{n \times m}$, $(\tilde \Xi^{-1}_m)^*=(\tilde \xi_1\cdots \tilde \xi_m)\in \Cr^{m \times m}.$ 
%However, the estimated eigen-modes  constitute potentially poor approximations. The lack of optimality of the algorithm originates from : 
%%  The quantities   computed in  Algorithm \ref{algo:00} are potentially  poor approximations of the  Koopman modes.
%In particular, we see that the approximation  is exact only if the  $\mu_j$'s  belong to the span of $\YYY$. Furthermore, we note that even in this case, the K-DMD algorithm relies on two important assumptions: {\it i)}  $\Psi^{-1}$ is  linear  so that model  \eqref{eq:eigenmoderepre} is  consistent, {\it ii)} operator $\AAA$ needs to be full-rank.  
%  the disregard for the low-rank constraint in the set $\mathcal{B}_k(\mathcal{H},\mathcal{H})$ in \eqref{eq:prob1}
 
The suitability of K-DMD  for  reduced modeling  is weak in the case $k < m$ and $p~\gg~1$.  Indeed, we can show that the computation burden to compute the approximation is independent of the reduced model dimension $k$, but requires  $m^2$ vector operations  in $\Rr^p$. More precisely,   the overall complexity of the K-DMD algorithm may be divided in an {\it on-line} and {\it off-line} part,  defined  as the  complexity  of the steps respectively independent or not of the  initial condition~$\theta$ we want to evaluate. We observe that the off-line part requires  $\mathcal{O}(m^2(m+p))$ operations to compute the matrix products in step 1),  the eigen-decompositions in steps 2) and 3) and the pseudo-inversion in step 4), while the on-line 
  part requires   $\mathcal{O}(pm^2)$ operations to compute  the least-square solution  in step 7).  Thus the on-line computation is lighter than the off-line one but the complexity is  in $\mathcal{O}(pm^2)$ whatever the reduced model dimension.  A consequence is that the lower $k$, the higher the ratio complexity/accuracy, which is obviously not a desirable property for reduced modeling.

\section{Proof of Proposition \ref{prop:3.3}}\label{app:3}\vspace{-0.1cm}
Let $$\tilde A_{\ell,k}=(R\,  \BBB^* \BBB\, S_k^*S_k \, \BBB^* \AAA\, R^*)^* \quad \textrm{and}\quad \tilde A_{r,k}= \quad  S_k\,  \BBB^* \BBB\, R^*\,R \, \AAA^* \BBB\, S_k^*.$$
%We will also use the closed-form solution $A^\star_k=\hat \R_k \hat \R_k^* \BBB \AAA^\dagger$ given in~\cite{HeasHerzet18Maps}.
We begin by proving  that the $U_{\AAA} \tilde \xi_i$'s are right eigen-vectors of $(A^\star_k)^*$.  
We  verify after some algebraic manipulations  that
$ U_{\AAA}^* (A^\star_k)^*  U_{\AAA}=(\tilde A_{\ell,k})^* .$ 
Then,  as    $U_{\AAA}$ is unitary we have
$(A^\star_k)^*  U_{\AAA} =U_{\AAA}(\tilde A_{\ell,k})^*, $  and in particular  $(A^\star_k)^*  U_{\AAA} \tilde \xi_i=U_{\AAA}(\tilde A_{\ell,k})^*  \tilde \xi_i.$
 Exploiting the fact that  $(\tilde \xi_i, \tilde \lambda_{i}) $ are eigen-vectors and eigen-values of $(\tilde A_{\ell,k})^*$, we obtain for $i=1,\ldots,k$ the sought result:   
$ (A^\star_k)^*  U_{\AAA} \tilde \xi_i=\tilde \lambda_{i} U_{\AAA} \tilde \xi_i.$

We continue by showing that the $\hat \R_k \tilde \zeta_i$'s are right eigen-vectors of $A^\star_k $. From the definition of $A^\star_k$ and $\tilde A_{r,k} $ we obtain
 $\hat \R_k^* A^\star_k \hat \R_k = \tilde A_{r,k}.$
Then, 
$ \hat \R_k \hat \R_k^*A^\star_k \hat \R_k  = \hat \R_k \tilde A_{r,k} ,$
which can be rewritten as 
$A^\star_k \hat \R_k  = \hat \R_k \tilde A_{r,k} ,$ since matrix $ \hat \R_k \hat \R_k^*$ is idempotent  and  $ \hat \R_k \hat \R_k^*A^\star_k=\hat \R_k \hat \R_k^*\hat \R_k \hat \R_k^* \BBB \AAA^\dagger=A^\star_k$. Because $\tilde \zeta_i$ are eigen-vectors of $ \tilde A_{r,k}$, we deduce for $i=1,\ldots,k$ the sought result:
$A^\star_k \hat \R_k \tilde \zeta_i = \tilde \lambda_{i} \hat \R_k  \tilde \zeta_i$. $\square$ \medskip

 \section{Inverse mapping}\label{app:1}
\subsection{Extended  definition}\label{app:1.1}
In this section, we first show that their exists an implicit definition of the inverse mapping for any element of the subset $\mathcal{S} =\{x\in \mathcal{H} : x=\Psi(y), y \in \Rr^p\}\subseteq \mathcal{H}$, where we recall that  $\mathcal{H}$ is a RKHS characterized by the reproducing kernel  \vspace{-0.cm}
\begin{align}\label{eq:kernelProp2}
\forall  (y,z) \in \Rr^p \times \Rr^p,\quad h(y,z)=\langle \Psi(y), \Psi(z)\rangle_\mathcal{H}.
\end{align}
We then propose an extension of this definition  for any element of $\mathcal{H}$.\\
 %,  we  impose  with this definition that the inverse of any elements  in $\mathcal{H}$  is associated to at least one element in $\Rr^p$. 
%In Section~\ref{sec:existence}, we discuss uniqueness and  existence of the reduced model approximation~\eqref{eq:approxROM}, which takes the form of the inverse mapping of a  linear combination of elements of $\mathcal{S}$. %Nevertheless, we show that the inverse is well defined over $\mathcal{H}$ for many kernel of interest. 
%  Finally, in Section~\ref{sec:approxInverse}, we provide closed-form approximations of this inverse in the case of polynomial or Gaussian kernels. 
 
First, let us show that the inverse of any $\eta \in \mathcal{S}$, $\Psi^{-1}(\eta)$  is implicitly defined as the unique element of $\Rr^p$ given by\vspace{-0.1cm}
\begin{align}\label{eq:defInverse}
\Psi^{-1}(\eta)=\argmax_{z \in  \Rr^p}\frac{\langle \eta,\Psi(z) \rangle_{\mathcal{H}}}{ \|\Psi(z)\|_{\mathcal{H}}}.
\end{align}
Indeed, the positive-definiteness of the kernel $h$ implies that we have for  any $y,z \in \Rr^p$
$$\det\begin{pmatrix} h(y,y) & h(y,z) \\ h(z,y) & h(z,z)\end{pmatrix} \ge 0,$$
 and  the lower bound is reached if and only if $y=z$. 
Using the kernel symmetry, this is equivalent to \vspace{-0.25cm}
\begin{align}
\frac{h(y,z)^2}{h(y,y)h(z,z)}& \le 1, \nonumber
\end{align}
and,  according to~\eqref{eq:kernelProp2}, it follows that 
\begin{align}\label{eq:kernelIneq}
 \frac{\langle \Psi(y),\Psi(z) \rangle_{\mathcal{H}}^2}{\|\Psi(y)\|_{\mathcal{H}}^2 \|\Psi(z)\|_{\mathcal{H}}^2}  \le 1,
\end{align}
for any $y,z \in \Rr^p$ where the  upper bound is reached if and only if $y=z$. \eqref{eq:kernelIneq} then implies definition \eqref{eq:defInverse}.

Then, as proposed in  \eqref{eq:defInverse2}, we extend   definition \eqref{eq:defInverse} holding for elements of 
$\mathcal{S}$ to any $\eta$ in $ \mathcal{H} $ (\ie    including elements outside  $\mathcal{S}$), as an  element $\Psi^{-1}(\eta)$ of $\Rr^p$ satisfying 
\begin{align}\label{eq:defInverse2_bis}
\Psi^{-1}(\eta) \in
\argmin_{z \in  \Rr^p}\|{ \eta}-{\Psi(z)}\|_{\mathcal{H}}.
\end{align}
%where $\textrm{dist}(\eta,\zeta)=\|\eta-\zeta\|_{\mathcal{H}}$ for $\eta,\zeta \in \mathcal{H}$ is the distance associated to the norm.
The definition  \eqref{eq:defInverse} coincides with  definition   \eqref{eq:defInverse2_bis} for normalized kernels (or more generally if ${\|\Psi(z)\|_{\mathcal{H}}}=\textrm{cte}$). 
%However,   it is not clear if this inverse exists outside of $\mathcal{S}$  or for non-normalized kernels.   This existence is discussed in Appendix~\ref{sec:existence} for linear combinations of elements of $\mathcal{S}$, which constitutes our case of interest.

\subsection{Proof of Proposition \ref{prop:inverse} }\label{app:1.2}

 We  first deduce from Proposition~\ref{prop:3.3} the closed-form $i$th   eigen-function approximation $ \varphi_i(\theta)$ for  $i=1,\ldots, k$  at any point  $\theta \in \Rr^p$ 
 \begin{align*}
 \varphi_i(\theta)=  \langle \xi_i, \Psi(\theta) \rangle_{\mathcal{H}}=  \langle U_{\AAA} \tilde \xi_i, \Psi(\theta) \rangle_{\mathcal{H}}= %(\Psi(\theta)^*  \AAA) R^* \tilde \xi_i
 \tilde \xi_i^* R \AAA^*  \Psi(\theta).\vspace{-0.2cm}
\end{align*}
Then, noticing that $$ \hat \R_k=\BBB  \mathbb{P}_ {\AAA^*}S_k^*=\BBB S_k^*,$$ as by construction $ \textrm{span}(S_k^*) \subseteq \textrm{span}(V_{\ZZZ}) \subseteq \textrm{span}( \mathbb{P}_ {\AAA^*})$,  relying again on Proposition~\ref{prop:3.3} we can rewrite the reduced model 
%$$\hat \R_k= \BBB \mathbb{P}_{\AAA^*}V_\ZZZ (\textrm{diag}\begin{pmatrix} \sigma_\ZZZ^1 & \cdots & \sigma_\ZZZ^k & 0 &\cdots \end{pmatrix})^\dagger,$$
%which implies that
 \eqref{eq:koopman1} in terms of  $\tilde \zeta_i$'s,  $\varphi_i(\theta)$'s and $\tilde \lambda_{i}$'s as\vspace{-0.1cm}
\begin{align}\label{eq:approxROM}
\tilde x_{\TT}(\theta)%&=\Psi^{-1}(\sum_i \zeta_i \lambda_i^{t-1} \varphi_i(\theta)),\nonumber \\
&=\Psi^{-1}(\sum_{j=1}^k \hat \R_k \tilde \zeta_j  \tilde \lambda_{j}^{{\TT}-1} \varphi_j(\theta)) 
%=\Psi^{-1}\left(\sum_{i=1,\TTPrim=1}^{N,T-1}\Psi(x_{\TTPrim+1}(\theta_i)) g^{\theta,\TT}_{(T-1)i+\TTPrim}\right)
=\Psi^{-1}(\BBB g^{\theta,{\TT}}),
\end{align}
with $ g^{\theta,{\TT}}$ defined in Proposition~\ref{prop:inverse}.
%\textrm{with}\quad g^{\theta,{\TT}}&=S_k^* (\tilde \zeta_1 \cdots \tilde \zeta_k)  \begin{pmatrix}\tilde \lambda^{{\TT}-1}_{\ell,1} \varphi_1(\theta) &\cdots& \tilde \lambda^{{\TT}-1}_{\ell,k}\varphi_k(\theta)\end{pmatrix}^* \in  \Rr^{m}.\label{eq:g}
%\vspace{-0.cm}
%\end{align}
Equation \eqref{eq:approxROM} implies the inverse of a linear combination of the $\Psi(y_i)$'s, where  $y_{i}=x_{t+1}(\thetaLearn_j)$ with $i=(\TTPrim-1)j+t$ for $j=1,\ldots, \NNPrim$ and $t=1,\ldots, \TTPrim-1$. From \eqref{eq:defInverse2}, we  rewrite the inverse of the linear combination appearing in \eqref{eq:approxROM} in terms of scalar products in $\mathcal{H}$ \vspace{-0.cm}
 \begin{align*}
\tilde x_{\TT}(\theta)&\in \arg\min_{z\in \Rr^p} \|{ \BBB g^{\theta,{\TT}}}-{\Psi(z)}\|_{\mathcal{H}} 
= \arg\min_{z\in \Rr^p} \|{ \BBB g^{\theta,{\TT}}}-{\Psi(z)}\|^2_{\mathcal{H}}  \nonumber\\
&= \arg\min_{z\in \Rr^p} \|{ {\Psi(z)}\|^2_{\mathcal{H}}-2\langle \BBB g^{\theta,{\TT}}},\Psi(z)\rangle_{\mathcal{H}}. \vspace{-0.cm} \nonumber
%&= \arg\min_{z\in \Rr^p} \left(  h(z,z)-2\sum_{i=1}^m  g_i^{\theta,{\TT}} {h(y_i,z)} \right). \vspace{-0.cm}%\label{eq:objectiveGeneral}
\end{align*}
The sought minimization problem is obtained by evaluating scalar products in $\mathcal{H}$ with the kernel trick $\square$.

\subsection{Sufficient conditions for existence}\label{sec:existence}\label{app:1.3}
 %Without loss of generality,  we assume a normalized kernel, \ie that $\|\Psi(z)\|_{\mathcal{H}}=1$  for any $\eta \in \mathcal{H}$.
In this section, we provide a set of mild sufficient conditions so that the inverse $\Psi^{-1}(\sum_i^m g_i^{\theta,\TT} \Psi(y_i))$ exists, $\Psi^{-1}$ being defined in \eqref{eq:defInverse2_bis}. 
This inverse is involved in  \eqref{eq:defInverse3} and was reformulated as the  problem of the existence of a vector $z^\star \in \Rr^p$ satisfying
$
f(z^\star)=\inf_{z\in \Rr^p}  f(z),
$
where the objective function is 
$
f(z)=h(z,z)-2\sum_{i=1}^m  g_i^{\theta,\TT} {h(y_i,z)}.$\vspace{-0.15cm}\\

%The  solution \eqref{eq:approxROM}  of our reduced modeling problem is the inverse of a linear combination $\eta=\sum_i^m g_i^{\theta,\TT} \Psi(y_i)$ of elements $\Psi(y_i)$ of $\mathcal{S}$. In the following proposition, we provide conditions guaranteeing the existence of the inverse  \eqref{eq:defInverse2_bis}, or equivalently, as done to obtain \eqref{eq:defInverse3}, that 
%\begin{align*}
%\|{\Psi(z^\star)}\|^2_{\mathcal{H}}-2\langle{ \eta},{\Psi(z^\star)}\rangle_{\mathcal{H}}& =
%\inf_{z\in \Rr^p}  f(z), %\label{eq:objectiveGeneral}
%\end{align*}

\begin{proposition}\label{prop:c1}
A vector $ \arg\inf_{z\in \Rr^p}  f(z)$ exists if one of the two following conditions holds.
\begin{itemize}
\item[(1)]  $f(z)$ is coercive and lower semicontinuous. 
\item[(2)]  \begin{itemize}
\item[i)] $h(x,x)$ is constant for any $x\in \Rr^p$,  
\item[ii)]$\lim_{\|z\|\to \infty} h(x,z)=0$ for  any $x\in \Rr^p$,
\item[iii)] $\int_{\Rr^p} h(x,z)dz=c$ with $c\ge 0$  constant for any $x\in \Rr^p$, 
\item[iv)] $f(z)$ is lower semicontinuous, 
\item[v)]  $\sum_{i=1}^m  g_i^{\theta,\TT} >0$.\\
\end{itemize}
\end{itemize}
\end{proposition}

{ 
{\it Proof of condition (1).} As  $\Rr^p$ is closed, the Weierstrass' Theorem \cite[Proposition A.8]{bertsekas1995nonlinear} states that  a minimizer exists  under condition {\it (1)}. $\square$\\
 
{\it Proof of condition (2).} Under condition {\it (2-i)},   we have \vspace{-0.1cm} 
\begin{align}\label{eq:arginf}
\arg\inf_{z\in \Rr^p} f(z)&=\arg\inf_{z\in \Rr^p} \left(-\sum_{i=1}^m  g_i^{\theta,\TT} h(y_i,z)\right), \vspace{-0.1cm} 
%&=\arg\inf_{z\in \Rr^p} \left( -\sum_{i=1}^m  g_i^{\theta,\TT} {e^{-\alpha\|y_i-z\|^\xi_\zeta_2}}-\sum_{i=1| g_i^{\theta,\TT}< 0}^m |g_i^{\theta,\TT}| \right), \nonumber \\
\end{align}
and condition {\it (2-ii)} implies that for $i=1\ldots,m$ we have \vspace{-0.1cm} 
\begin{align}\label{eq:tend0}
\lim_{\|z\|\to \infty} -\sum_{i=1}^m  g_i^{\theta,\TT} h(y_i,z)=0.\vspace{-0.1cm} 
\end{align}
%\begin{lemma}\label{lemmac2}
Moreover, under conditions {\it (2-iii)} and {\it (2-v)}, \vspace{-0.1cm} 
\begin{align}\label{lemmac2}
\exists z^\star \in \Rr^p \quad \textrm{such\,that}\quad -\sum_{i=1}^m  g_i^{\theta,\TT} {h(y_i,z^\star)}<0.\vspace{-0.1cm} 
\end{align}
%\end{lemma}
%\proof{
Indeed, by contraposition, assume that $ -\sum_{i=1}^m  g_i^{\theta,\TT} {h(y_i,z)}\ge 0$ for all $z\in \Rr^p$. By integrating with respect to $z$, as each term in the sum is independent of the point  $y_i$ under condition {\it (2-iii)}, we obtain that $-c\sum_{i=1}^m  g_i^{\theta,\TT} \le 0$ and as $c\ge 0$, that $\sum_{i=1}^m  g_i^{\theta,\TT}\le 0$, which contradicts  {\it (2-v)}. 
%}

The existence  of a minimizer then  follows from  \eqref{eq:arginf}, \eqref{eq:tend0}, \eqref{lemmac2} and assumption~{\it (2-iv)}. Indeed,  assume a sequence $\{x_k\}\subset \Rr^p$ such that $\lim_{k \to \infty} f(x_k)=\inf_{z \in \Rr^p}f(z)$. $\{x_k\}$ must be bounded  since \eqref{eq:arginf} and \eqref{eq:tend0} show, that on the one hand the objective function $f(z)$ tends to $0$ as the norm of $z$ grows to infinity, and on the other hand~\eqref{lemmac2} guarantees that there exists at least one point for which the objective function is negative. Using the lower semicontinuity of $f$, the proof then proceeds like the proof of the Weierstrass' Theorem \cite[Proposition A.8]{bertsekas1995nonlinear}. $\square$

% and the logarithm in \eqref{eq:objective} is well-defined since  we have that
%\begin{align*}
% \sum_{i=1}^m  g_i^{\theta,\TT} e^{-\alpha\|y_i-z\|^\xi_\zeta}+\sum_{i=1| g_i^{\theta,\TT}< 0}^m|g_i^{\theta,\TT}|  &\ge  \sum_{i=1}^m  g_i^{\theta,\TT} e^{-\alpha\|y_i-z\|^\xi_\zeta}+\sum_{i=1| g_i^{\theta,\TT}< 0}^m|g_i^{\theta,\TT}| e^{-\alpha\|y_i-z\|^\xi_\zeta} \\
% &=  \sum_{i=1 | g_i^{\theta,\TT}\ge 0}^m  g_i^{\theta,\TT} e^{-\alpha\|y_i-z\|^\xi_\zeta}  \ge 0.
%\end{align*}
% letting the magnitude of any component $z_j$ of $z$ grow to infinity, we have
%\begin{align}\label{eq:cond1}
%\lim_{|z_j| \to \infty} -\log\left( \sum_{i=1}^m  g_i^{\theta,\TT} e^{-\alpha\|y_i-z\|^\xi_\zeta}\right)_+&=+\infty.
%\end{align}

}
\subsection{Examples}\label{sec:examples}
The sufficient conditions for existence of the inverse hold for several class of kernels.
In particular, condition (1)  holds for the polynomial kernels of the form
$ h(y_i,z)=(1+y_i^* z)^\gamma,\quad \gamma \in \mathbb{N}^*.$
 Indeed,  the objective function
$f(z)= (1+\|z\|_2^2)^\gamma - 2 \sum_i^m g_i^{\theta,\TT}(1+ y_i^*z)^\gamma%\\
$ is in this case continuous and  coercive since 
$
 \lim_{\|z\| \to \infty} f(z) = \lim_{\|z\| \to \infty} \|z\|_2^{2\gamma} = +\infty.%\left(1-2\sum_i^m g_i^{\theta,\TT} \|y_i\|_2^\gamma(1-\epsilon_i)^\gamma\right) 
$
Conditions {\it (2-i)},  {\it (2-ii)},  {\it (2-iii)}  and  {\it (2-iv)} are in particular verified  for  Gaussian kernels of the form 
 $h(y_i,z)=\exp^{-\frac{1}{2\sigma^{2}}\|y_i- z\|_2^2}$ with $\sigma >0,$
or    Laplacian kernels  of the form
 $h(y_i,z)=\exp^{-\frac{1}{\beta} \|y_i- z\|_2}$ with $\beta >0.$
The existence of the inverse is therefore guaranteed for these kernels as long as we check that condition {\it (2-v)} holds. This positivity condition was always observed in our numerical simulations.\medskip

  \textbf{Acknowledgements.}
The  authors  would  like to  thank  Mathias  Rousset  for his help in proving~Proposition~\ref{prop:c1}. \\\vspace{-0.cm}

 \textbf{Conflicts of interest and data availability}
The authors have no conflicts of interest to declare that are relevant to the content of this article.  The data presented in this manuscript is owned by the authors and no permissions are required. \vspace{-0.2cm}

	 \bibliographystyle{spmpsci}     
\bibliography{./bibtex}

\begin{thebibliography}{10}
\providecommand{\url}[1]{{#1}}
\providecommand{\urlprefix}{URL }
\expandafter\ifx\csname urlstyle\endcsname\relax
  \providecommand{\doi}[1]{DOI~\discretionary{}{}{}#1}\else
  \providecommand{\doi}{DOI~\discretionary{}{}{}\begingroup
  \urlstyle{rm}\Url}\fi

\bibitem{alexander2020operator}
Alexander, R., Giannakis, D.: Operator-theoretic framework for forecasting
  nonlinear time series with kernel analog techniques.
\newblock Physica D: Nonlinear Phenomena \textbf{409}, 132,520 (2020)

\bibitem{bach2005predictive}
Bach, F.R., Jordan, M.I.: Predictive low-rank decomposition for kernel methods.
\newblock In: Proceedings of the 22nd international conference on Machine
  learning, pp. 33--40 (2005)

\bibitem{benner2020operator}
Benner, P., Goyal, P., Kramer, B., Peherstorfer, B., Willcox, K.: Operator
  inference for non-intrusive model reduction of systems with non-polynomial
  nonlinear terms.
\newblock Computer Methods in Applied Mechanics and Engineering \textbf{372},
  113,433 (2020)

\bibitem{bertsekas1995nonlinear}
Bertsekas, D.: Nonlinear Programming.
\newblock Athena Scientific (1995)

\bibitem{billaud2017dynamical}
Billaud-Friess, M., Nouy, A.: Dynamical model reduction method for solving
  parameter-dependent dynamical systems.
\newblock SIAM Journal on Scientific Computing \textbf{39}(4), A1766--A1792
  (2017)

\bibitem{bishop2006pattern}
Bishop, C.M.: Pattern Recognition and Machine Learning (Information Science and
  Statistics).
\newblock Springer-Verlag, Berlin, Heidelberg (2006)

\bibitem{bouvrie2017kernel}
Bouvrie, J., Hamzi, B.: Kernel methods for the approximation of nonlinear
  systems.
\newblock SIAM Journal on Control and Optimization \textbf{55}(4), 2460--2492
  (2017)

\bibitem{cagniart2019model}
Cagniart, N., Maday, Y., Stamm, B.: Model order reduction for problems with
  large convection effects.
\newblock Contributions to partial differential equations and applications pp.
  131--150 (2019)

\bibitem{chandrasekhar2013hydrodynamic}
Chandrasekhar, S.: Hydrodynamic and hydromagnetic stability.
\newblock Courier Corporation (2013)

\bibitem{Chen12}
Chen, K.K., Tu, J.H., Rowley, C.W.: Variants of dynamic mode decomposition:
  boundary condition, koopman, and fourier analyses.
\newblock Journal of nonlinear science \textbf{22}(6), 887--915 (2012)

\bibitem{das2021reproducing}
Das, S., Giannakis, D., Slawinska, J.: Reproducing kernel hilbert space
  compactification of unitary evolution groups.
\newblock Applied and Computational Harmonic Analysis \textbf{54}, 75--136
  (2021)

\bibitem{giannakis2023learning}
Giannakis, D., Henriksen, A., Tropp, J.A., Ward, R.: Learning to forecast
  dynamical systems from streaming data.
\newblock SIAM Journal on Applied Dynamical Systems \textbf{22}(2), 527--558
  (2023)

\bibitem{golub2013matrix}
Golub, G., Van~Loan, C.: Matrix Computations.
\newblock Johns Hopkins Studies in the Mathematical Sciences. Johns Hopkins
  University Press (2013)

\bibitem{heas2017optimal}
H{\'e}as, P., Herzet, C.: Optimal low-rank dynamic mode decomposition.
\newblock In: 2017 IEEE International Conference on Acoustics, Speech and
  Signal Processing (ICASSP), pp. 4456--4460. IEEE (2017)

\bibitem{HeasHerzet18Maps}
H\'eas, P., Herzet, C.: Low-rank approximation of linear maps.
\newblock arXiv e-prints  (2018)

\bibitem{HeasHerzet17}
H{\'e}as, P., Herzet, C.: {Low-Rank Dynamic Mode Decomposition: An Exact and
  Tractable Solution}.
\newblock {Journal of Nonlinear Science} \textbf{32}(1) (2022)

\bibitem{HeasIcassp2020}
{H\'eas}, P., {Herzet}, C., {Comb\`es}, B.: Generalized kernel-based dynamic
  mode decomposition.
\newblock In: IEEE International Conference on Acoustics, Speech and Signal
  Processing (ICASSP) (2020)

\bibitem{heas2014self}
H{\'e}as, P., Lavancier, F., Kadri-Harouna, S.: Self-similar prior and wavelet
  bases for hidden incompressible turbulent motion.
\newblock SIAM Journal on Imaging Sciences \textbf{7}(2), 1171--1209 (2014)

\bibitem{hemati2017biasing}
Hemati, M.S., Rowley, C.W., Deem, E.A., Cattafesta, L.N.: De-biasing the
  dynamic mode decomposition for applied {K}oopman spectral analysis of noisy
  datasets.
\newblock Theoretical and Computational Fluid Dynamics \textbf{31}(4), 349--368
  (2017)

\bibitem{Horn12}
Horn, R.A., Johnson, C.R.: Matrix analysis.
\newblock Cambridge university press (2012)

\bibitem{iollo2014advection}
Iollo, A., Lombardi, D.: Advection modes by optimal mass transfer.
\newblock Physical Review E \textbf{89}(2), 022,923 (2014)

\bibitem{Jovanovic12}
Jovanovic, M., Schmid, P., Nichols, J.: Low-rank and sparse dynamic mode
  decomposition.
\newblock Center for Turbulence Research Annual Research Briefs pp. 139--152
  (2012)

\bibitem{kawahara2016dynamic}
Kawahara, Y.: Dynamic mode decomposition with reproducing kernels for koopman
  spectral analysis.
\newblock Advances in neural information processing systems \textbf{29} (2016)

\bibitem{klus2020eigendecompositions}
Klus, S., Schuster, I., Muandet, K.: Eigendecompositions of transfer operators
  in reproducing kernel hilbert spaces.
\newblock Journal of Nonlinear Science \textbf{30}(1), 283--315 (2020)

\bibitem{koch2007dynamical}
Koch, O., Lubich, C.: Dynamical low-rank approximation.
\newblock SIAM Journal on Matrix Analysis and Applications \textbf{29}(2),
  434--454 (2007)

\bibitem{koopman1931hamiltonian}
Koopman, B.O.: Hamiltonian systems and transformation in hilbert space.
\newblock Proceedings of the National Academy of Sciences of the United States
  of America \textbf{17}(5), 315 (1931)

\bibitem{korda2018convergence}
Korda, M., Mezi{\'c}, I.: On convergence of extended dynamic mode decomposition
  to the koopman operator.
\newblock Journal of Nonlinear Science \textbf{28}(2), 687--710 (2018)

\bibitem{kowalski1991nonlinear}
Kowalski, K., Steeb, W.H.: Nonlinear dynamical systems and Carleman
  linearization.
\newblock World Scientific (1991)

\bibitem{Lorenz63}
{Lorenz}, E.N.: {Deterministic Nonperiodic Flow.}
\newblock Journal of Atmospheric Sciences \textbf{20}, 130--148 (1963)

\bibitem{Lusch2018DeepLF}
Lusch, B., Kutz, J.N., Brunton, S.L.: Deep learning for universal linear
  embeddings of nonlinear dynamics.
\newblock In: Nature Communications (2018)

\bibitem{mezic2004comparison}
Mezi{\'c}, I., Banaszuk, A.: Comparison of systems with complex behavior.
\newblock Physica D: Nonlinear Phenomena \textbf{197}(1-2), 101--133 (2004)

\bibitem{nocedal2000numerical}
Nocedal, J., Wright, S.: Numerical Optimization.
\newblock Springer Series in Operations Research and Financial Engineering.
  Springer New York (2000)

\bibitem{nouy2010proper}
Nouy, A.: Proper generalized decompositions and separated representations for
  the numerical solution of high dimensional stochastic problems.
\newblock Archives of Computational Methods in Engineering \textbf{17}(4),
  403--434 (2010)

\bibitem{peherstorfer2016data}
Peherstorfer, B., Willcox, K.: Data-driven operator inference for nonintrusive
  projection-based model reduction.
\newblock Computer Methods in Applied Mechanics and Engineering \textbf{306},
  196--215 (2016)

\bibitem{quarteroni2015reduced}
Quarteroni, A., Manzoni, A., Negri, F.: Reduced basis methods for partial
  differential equations: an introduction, vol.~92.
\newblock Springer (2015)

\bibitem{Schmid10}
Schmid, P.J.: Dynamic mode decomposition of numerical and experimental data.
\newblock Journal of Fluid Mechanics \textbf{656}, 5--28 (2010)

\bibitem{scholkopf1998nonlinear}
Sch{\"o}lkopf, B., Smola, A., M{\"u}ller, K.R.: Nonlinear component analysis as
  a kernel eigenvalue problem.
\newblock Neural computation \textbf{10}(5), 1299--1319 (1998)

\bibitem{steinwart2006explicit}
Steinwart, I., Hush, D., Scovel, C.: An explicit description of the reproducing
  kernel hilbert spaces of gaussian rbf kernels.
\newblock IEEE Transactions on Information Theory \textbf{52}(10), 4635--4643
  (2006)

\bibitem{taddei2020registration}
Taddei, T.: A registration method for model order reduction: data compression
  and geometry reduction.
\newblock SIAM Journal on Scientific Computing \textbf{42}(2), A997--A1027
  (2020)

\bibitem{tropp2023randomized}
Tropp, J.A., Webber, R.J.: Randomized algorithms for low-rank matrix
  approximation: Design, analysis, and applications.
\newblock arXiv preprint arXiv:2306.12418  (2023)

\bibitem{Tu2014391}
Tu, J.H., Rowley, C.W., Luchtenburg, D.M., Brunton, S.L., Kutz, J.N.: On
  dynamic mode decomposition: Theory and applications.
\newblock Journal of Computational Dynamics \textbf{1}(2), 391--421 (2014)

\bibitem{williams2015data}
Williams, M.O., Kevrekidis, I., Rowley, C.: A data--driven approximation of the
  koopman operator: Extending dynamic mode decomposition.
\newblock Journal of Nonlinear Science \textbf{25}(6), 1307--1346 (2015)

\bibitem{williams2014kernel}
Williams, M.O., Rowley, C.W., Kevrekidis, I.G.: A kernel-based method for
  data-driven koopman spectral analysis.
\newblock Journal of Computational Dynamics \textbf{2}(2), 247--265 (2015)

\bibitem{yang2012nystrom}
Yang, T., Li, Y.F., Mahdavi, M., Jin, R., Zhou, Z.H.: Nystr{\"o}m method vs
  random fourier features: A theoretical and empirical comparison.
\newblock Advances in neural information processing systems \textbf{25} (2012)

\bibitem{yeung2019learning}
Yeung, E., Kundu, S., Hodas, N.: Learning deep neural network representations
  for koopman operators of nonlinear dynamical systems.
\newblock In: 2019 American Control Conference (ACC), pp. 4832--4839. IEEE
  (2019)

\bibitem{zhuoperator}
Zhu, K.: Operator Theory in Function Spaces, Second Edition.
\newblock Mathematical surveys and monographs. American Mathematical Soc.
  (2007)

\end{thebibliography}
\end{document}